\documentclass{article} 
\usepackage{iclr2025_conference,times}

\usepackage{amsmath,amsfonts,bm}

\def\eqref#1{equation~\ref{#1}}

\def\1{\bm{1}}

\DeclareMathAlphabet{\mathsfit}{\encodingdefault}{\sfdefault}{m}{sl}
\SetMathAlphabet{\mathsfit}{bold}{\encodingdefault}{\sfdefault}{bx}{n}

\usepackage{hyperref}
\usepackage{url}
\usepackage{graphicx}

\usepackage{listings}
\usepackage{microtype}
\usepackage{algorithm}
\usepackage{algpseudocode}
\usepackage{amsthm}
\usepackage{booktabs} 
\usepackage{tabularx} 
\usepackage{arydshln} 
\usepackage{multirow} 
\usepackage{wrapfig} 
\usepackage{caption}
\usepackage{subcaption}
\usepackage{wrapfig}
\usepackage{enumitem}

\definecolor{Gray}{gray}{0.9}
\definecolor{LightCyan}{rgb}{0.88,1,1}
\definecolor{red}{RGB}{144,0,32}
\definecolor{blue}{rgb}{0.06, 0.3, 0.57}
\definecolor{warmblack}{rgb}{0.0, 0.26, 0.26}
\definecolor{purple}{rgb}{0.4, 0.01, 0.24}
\definecolor{tawny}{rgb}{0.8, 0.34, 0.0}
\definecolor{amber}{rgb}{1.0, 0.49, 0.0}
\definecolor{antiquefuchsia}{rgb}{0.57, 0.36, 0.51}
\usepackage{hyperref}
\usepackage{url}
\hypersetup{
    colorlinks=true,
    citecolor=antiquefuchsia,
    linkcolor=warmblack,
    urlcolor=purple,
    }

\title{Instruction-Guided Autoregressive Neural Network Parameter Generation}

\newcommand*{\affaddr}[1]{#1}
\newcommand*{\affmark}[1][*]{\textsuperscript{#1}}
\newcommand*{\email}[1]{\texttt{#1}}
\newcommand*\samethanks[1][\value{footnote}]{\footnotemark[#1]}

\author{%
Soro Bedionita\affmark[1]\thanks{Equal contribution.}, %
Bruno Andreis\affmark[1]\samethanks, %
Song Chong\affmark[1], %
Sung Ju Hwang\affmark[1,2] \\[6pt]
\affaddr{\affmark[1]KAIST AI}, 
\affaddr{\affmark[2]DeepAuto.ai}, South Korea \\[3pt]
\email{\{sorobedio,andries,songchong, sungju.hwang\}@kaist.ac.kr}
}

\newcommand{\ourmethod}{IGPG}
\newtheorem{remark}{Remark}

\iclrfinalcopy 
\begin{document}

\maketitle
\begin{abstract}
Learning to generate neural network parameters conditioned on task descriptions and architecture specifications is pivotal for advancing model adaptability and transfer learning. Existing methods—especially those based on diffusion models—suffer from limited scalability to large architectures, rigidity in handling varying network depths, and disjointed parameter generation that undermines inter-layer coherence. In this work, we propose IGPG (Instruction-Guided Parameter Generation), an autoregressive framework that unifies parameter synthesis across diverse tasks and architectures.
IGPG leverages a VQ-VAE and an autoregressive model to generate neural network parameters, conditioned on task instructions, dataset, and architecture details. By autoregressively generating neural network weights' tokens, IGPG ensures inter-layer coherence and enables efficient adaptation across models and datasets. Operating at the token level, IGPG effectively captures complex parameter distributions aggregated from a broad spectrum of pretrained models.
Extensive experiments on multiple vision datasets demonstrate that IGPG consolidates diverse pretrained models into a single, flexible generative framework. The synthesized parameters achieve competitive or superior performance relative to state-of-the-art methods, especially in terms of scalability and efficiency when applied to large architectures. These results underscore IGPG’s potential as a powerful tool for pretrained weight retrieval, model selection, and rapid task-specific fine-tuning.
\end{abstract}

\section{Introduction}
\label{sec:intro}

Deep neural networks have driven breakthroughs across domains—from image recognition~\citep{ILSVRC15, He2016DeepRL} to natural language processing—leading to vast repositories of pretrained models~\citep{schurholt2022model} available via platforms like Hugging Face\footnote{\url{https://huggingface.co/}} and libraries such as TIMM~\citep{rw2019timm}. Despite their success, adapting these models to new tasks or datasets is challenging. It often requires manual intervention, extensive fine-tuning, and careful model selection.

Prior work in transfer learning, meta-learning, and knowledge distillation~\citep{Gou-2021,Yang2021ASO,Elsken2019NeuralAS} has predominantly focused on individual models, often overlooking the cross-task insights embedded in large-scale model collections. More recent efforts in hyper-representation learning~\citep{schurholt-self-supervised-2021, schrholt2022hyperrepresentation, Schrholt2022HyperRepresentationsAG, wang2024neural} have sought to learn distributions over network weights to enhance initialization. However, these methods are typically unconditional and limited to single-task scenarios, neglecting the potential benefits of incorporating pretrained dataset embeddings during training. While a few studies have explored task-specific parameter generation~\citep{soro2024diffusionbased}, there remains a significant gap in developing a unified and flexible solution. Furthermore, when applied to large architectures, these approaches tend to generate weight chunks without considering the relationships within each sampled layer, thereby limiting performance and increasing sampling time.

To address these challenges, we introduce Instruction-Guided Parameter Generation (IGPG), a novel framework that integrates Vector Quantized Variational Autoencoders (VQ-VAE) with autoregressive modeling to generate neural network parameters conditioned on both task and architecture. IGPG jointly encodes three key elements: 
\textbf{Task Representations}: Using dataset embeddings or natural language instructions to capture target task semantics; \textbf{Architecture Specifications}: Explicitly representing network designs to enable cross-architecture parameter generation; and \textbf{Inter-Layer Dependencies}: Employing autoregressive modeling to preserve coherence across layers. This joint conditioning enables IGPG to efficiently synthesize coherent, task-optimized parameters, reducing reliance on extensive fine-tuning.

\begin{figure}[t!]
\begin{center}
\centerline{\includegraphics[width=0.7\columnwidth]{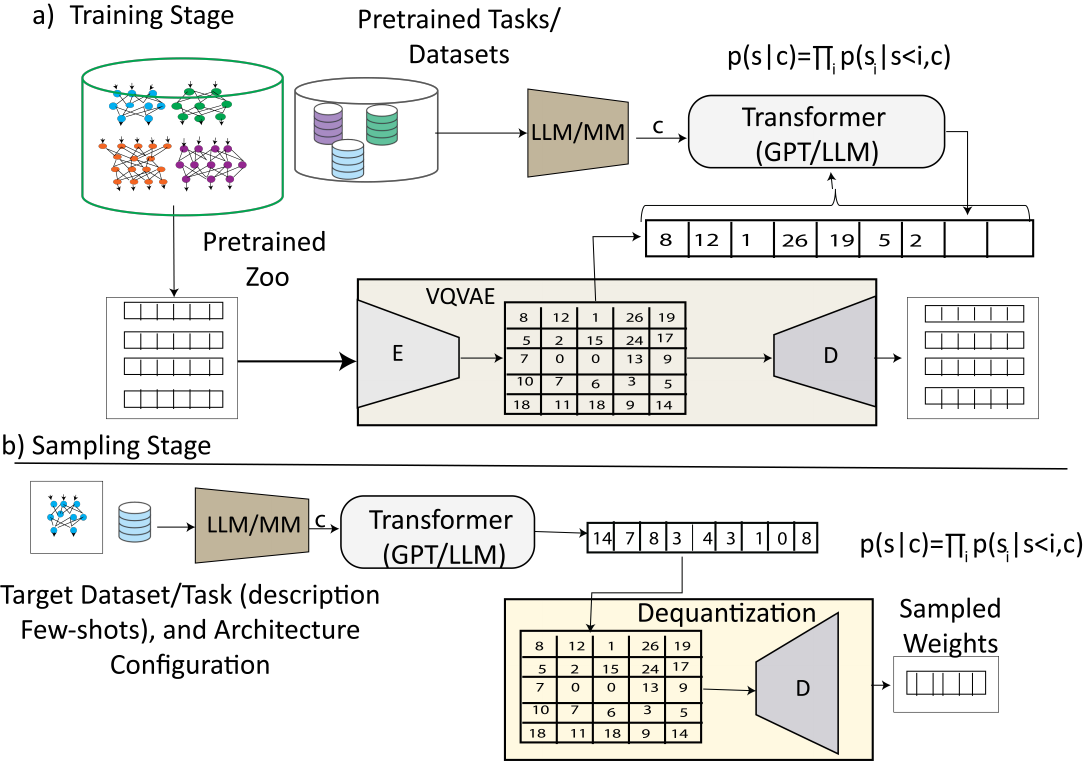}}
\caption{Our approach integrates a VQ-VAE autoencoder (\(\mathbf{E}\)–\(\mathbf{D}\)) with a transformer prior. First, the VQ-VAE encodes vectorized network parameters (see Section \ref{approach:enc}), and then the transformer is trained on the resulting codebook (see Section \ref{sec:autoreg}). Additionally, prompts—including data, task, or architecture details—are processed using multimodal or language modeling techniques (see Section \ref{sec:autoreg}), with an example training simplified prompt template provided in Remark \ref{tmp}.}
\label{overw}
\end{center}
\vskip -0.20in
\end{figure}

Our contributions are summarized as follows:
\begin{enumerate}[itemsep=0pt, topsep=0pt, parsep=0pt, partopsep=0pt]
    \item \textbf{Task-Conditioned Generation}: We propose a mechanism for directly generating network parameters from natural language or dataset descriptors, offering intuitive task control.
    \item \textbf{Architecture-Agnostic Framework}: Our method generates parameters across diverse architectures, leveraging knowledge from multiple pretrained models.
    \item \textbf{Autoregressive Coherence}: By modeling layer-wise dependencies, IGPG ensures internally consistent parameters that accelerate convergence and enhance transfer performance.
\end{enumerate}
Extensive experiments demonstrate that IGPG compresses and transfers the collective knowledge of diverse pretrained models into a single generative framework, achieving competitive or superior performance on unseen tasks and scaling effectively to larger architectures (see Figure~\ref{overw}).
\section{Instruction-Guided Parameters Generation}
\label{sec:formatting}
\subsection{Preliminary}

We introduce \emph{Instruction-Guided Parameter Generation} (IGPG), a framework that learns the distribution of pretrained models to generate new weights on demand (see Figure~\ref{overw}). By capturing key feature patterns of high-performing networks, IGPG can generate specialized weights for both existing and novel tasks or dataset, reducing extensive retraining and accelerating model deployment in diverse computer vision scenarios.

Our method begins with a set of pretrained models \(\{\theta_i\}_{i=1}^N\) and their corresponding datasets or task description \(\{D_i\}_{i=1}^N\). We construct a \emph{model zoo} by vectorizing each network’s parameters in one of two ways. In the \emph{layer-wise} setting, each layer’s weights (including biases) are flattened into a single vector, yielding per-layer parameter samples from across all pretrained networks. Alternatively, in the \emph{architecture-wise} setting, the flattened layer weights of each model are sequentially concatenated to form a single global parameter vector per model, preserving the original layer order. Both approaches produce uniform parameter representations that IGPG uses to learn a generative mapping, enabling the generation of dataset/task- and architecture-specific weights for efficient adaptation.

We formalize our setup by defining \(\mathcal{D}\) as the space of possible datasets or tasks, \(\mathcal{A}\) as the space of neural architectures, and \(\Theta\) as the parameter space. Our generative mapping \(H\) operates in two phases: during training, \(H:\mathcal{D}\times \mathcal{A}\times \Theta \rightarrow \Theta\); during inference, \(H:\mathcal{D}\times \mathcal{A}\rightarrow \Theta\). Thus, given a dataset \(D_i\) and architecture \(A_i\), the trained \(H\) produces a tailored initialization \(\hat{\theta}_i = H(D_i,A_i)\). To enforce autoregressive parameter generation and capture layer-wise dependencies, we build IGPG  based on a VQGAN structure combined with a transformer autoregressive prior. This design ensures coherent parameter generation by modeling dependencies between layers while leveraging the strengths of both architectures.

\subsection{Neural Network Parameters Encoding with VQVAE}\label{approach:enc}

We encode neural network parameters using a Gumbel Vector Quantized Variational Autoencoder (VQVAE)~\citep{Oord2017NeuralDR} to generate discrete representations suitable for autoregressive modeling. For a parameter vector $\Theta \in \mathbb{R}^D$, we employ fixed-size chunking with chunk size $K$, padding $\Theta$ to length $D' = \lceil \frac{D}{K} \rceil \times K$ and splitting it into $n = \frac{D'}{K}$ chunks for efficient processing.

The VQVAE architecture consists of an encoder $\mathbf{E}$, decoder $\mathbf{D}$, and quantization module $\mathbf{Q}$ with codebook $\mathbf{e} = \{e_1, ..., e_m\}$. For input parameters $\theta$, the encoder produces latent representations $z = \mathbf{E}(\theta)$, which are quantized using Gumbel-Softmax sampling:

\begin{equation}
    z_q = \sum_{j=1}^{m} y_j e_j, \quad y_j = \frac{\exp((\log \pi_j + g_j) / \tau)}{\sum_{i=1}^{m} \exp((\log \pi_i + g_i) / \tau)}
    \label{eqn:gqant}
\end{equation}

where $\pi_j$ are encoder logits, $g_j$ are Gumbel noise samples, and $\tau$ is the temperature parameter. The decoder reconstructs the input as $\hat{\theta} = \mathbf{D}(z_q)$. The model is optimized by minimizing:

\begin{equation}
    \mathcal{L} = \underbrace{\|\mathcal{M} \odot (\theta - \hat{\theta})\|_2^2}_{\text{reconstruction}} + \gamma \underbrace{\|z - \text{sg}[z_q]\|_2^2}_{\text{commitment}} + \beta \underbrace{\|\text{sg}[z] - z_q\|_2^2}_{\text{codebook}}
    \label{eqn:vqvae_loss}
\end{equation}
where $\mathcal{M}$ masks padding values, $\text{sg}[\cdot]$ denotes stop-gradient, and $\{\beta, \gamma\}$ are balancing coefficients. This Gumbel-VQVAE formulation enables stochastic, differentiable quantization while preparing parameter vectors for subsequent autoregressive modeling using transformer architectures.
\subsection{Autoregressive Modeling of Encoded Parameters}
\label{sec:autoregressive}
We design an autoregressive framework that conditions parameter generation on both dataset content and network architecture. For a labeled dataset \(\mathcal{D}\), we sample a balanced subset (e.g., five images per class) and embed each image using CLIP~\citep{radford2021learningtransferablevisualmodels}. Mean-pooling these embeddings yields the dataset-level vector \(\bar{e}_\mathcal{D}\). To encode a network architecture \(\mathcal{A}\), we convert its specifications into a standardized textual description \(\mathrm{desc}(\mathcal{A})\) and process it with LLaMA-3-Instruct~\citep{dubey2024llama3herdmodels}, producing the architecture-level embedding \(e_\mathcal{A}\).
In the transformer’s forward pass, we concatenate \(\bar{e}_\mathcal{D}\), \(e_\mathcal{A}\), and the VQVAE codebook embeddings to form a unified representation for conditioning the autoregressive prior (e.g., GPT-2).

\textbf{Training Process}: Following VQGAN~\citep{esser2020taming}, we employ a transformer-based prior (mini-GPT~\citep{radford2019language}) for conditional sampling. For each pretrained model, its encoded tokens are gathered into a single sequence representing the network. Our VQVAE is structured so that during training the autoregressive model can generate the full codebook in one pass based on next tokens prediction procedure where the context length is the length of the larger sequence vector. We train the GPT-based transformer to model the sequence likelihood and minimize the corresponding loss in a single formulation:
\begin{equation}
    \mathcal{L}_{\text{prior}} = \mathbb{E}_{s \sim p(s,e_\mathcal{A},\bar{e}_\mathcal{D})}\left[\log p(s \mid e_\mathcal{A},\bar{e}_\mathcal{D})\right], \quad \text{where} \quad p(s \mid e_\mathcal{A},\bar{e}_\mathcal{D}) = \prod_{i} p(s_i \mid s_{<i}, e_\mathcal{A},\bar{e}_\mathcal{D}).
    \label{eq:combined}
\end{equation}
Equation~\eqref{eq:combined} encapsulates our training objective, aligning generated parameter tokens with both dataset and architectural embeddings for coherent and efficient parameter synthesis. 

\section{Autoregressive Parameter Generation}
\label{sec:autoreg}

We consider a target architecture $\mathcal{A}$ whose parameters are given by a vector 
$\theta_{\mathcal{A}} \in \mathbb{R}^L$. To represent $\theta_{\mathcal{A}}$ with a 
manageable token sequence, we split it into $
  k = \left\lceil \frac{L}{K} \right\rceil$ 
chunks, each of size $K$. A learned VQ-VAE tokenizer $\mathcal{T}$ maps each chunk 
from $\mathbb{R}^K$ to a sequence of $l$ tokens from a discrete codebook $\mathcal{V}$, 
i.e.\ $\mathcal{T}: \mathbb{R}^K \rightarrow \mathcal{V}^l$. Consequently, the entire 
parameter vector $\theta_{\mathcal{A}}$ can be expressed via a token sequence of length 
$kl$. A VQ-VAE decoder $\mathbf{D}$ recovers real-valued parameter chunks from these tokens, 
$\mathbf{D}: \mathcal{V}^l \rightarrow \mathbb{R}^K$, and a flattening operator 
$\mathcal{F}$ reassembles the $k$ decoded chunks into the full parameter vector 
$\theta_{\mathcal{A}}$. Given a maximum token-sequence length $N_{\text{max}}$ observed in training, we distinguish 
two modes of parameter generation. In the simpler scenario where $kl \le N_{\text{max}}$, 
a single-stage procedure suffices. Specifically, an autoregressive model $\mathcal{H}$ 
takes as input an architecture embedding $e_{\mathcal{A}}$ and a dataset/task embedding 
$e_{\mathcal{D}}$ --- collectively denoted by 
$(e_{\mathcal{A}}, e_{\mathcal{D}})$ --- and outputs a sequence 
$\mathbf{s} \in \mathcal{V}^{kl}$. Splitting $\mathbf{s}$ into $k$ segments 
$\mathbf{s}_1, \dots, \mathbf{s}_k$, each of length $l$, and decoding them with 
$\mathbf{D}$ reconstructs the $k$ parameter chunks. Finally, flattening these chunks 
with $\mathcal{F}$ produces $\theta_{\mathcal{A}}$. For larger architectures, where $kl > N_{\text{max}}$, we adopt chunk-wise autoregressive 
generation. Here, the model cannot generate all $kl$ tokens at once without exceeding 
its maximum context size. Instead, we first generate an initial sequence 
$\mathbf{s}^{(1)} \in \mathcal{V}^{N_{\text{max}}}$ via 
$\mathcal{G}(e_{\mathcal{A}}, e_{\mathcal{D}})$. We then iteratively generate additional 
token blocks $\mathbf{s}^{(2)}, \dots, \mathbf{s}^{(J)}$, each conditioned on 
$(e_{\mathcal{A}}, e_{\mathcal{D}})$ and a context window from the previously generated 
block, where $  J = \left\lceil \frac{kl}{N_{\text{max}}} \right\rceil.
$
Concatenating all blocks yields 
$\mathbf{s}_{\text{full}} \in \mathcal{V}^{N_{\text{max}} \cdot J}$, which we truncate 
to the first $kl$ tokens if necessary. Finally, we split $\mathbf{s}_{\text{full}}$ 
into $k$ segments of length $l$ and decode each via $\mathbf{D}$ to form the $k$ 
chunks in $\mathbb{R}^K$. Flattening these chunks with $\mathcal{F}$ produces the full 
parameter vector $\theta_{\mathcal{A}}$. Thus, by leveraging a chunked VQ-VAE representation and limiting each generation step 
to $N_{\text{max}}$ tokens, we enable parameter generation for arbitrarily large 
architectures. Whenever $kl \le N_{\text{max}}$, a single-step generation suffices; 
otherwise, we compose multiple chunks autoregressively. This design efficiently scales 
the generation process while maintaining the model’s capacity to represent 
high-dimensional parameter vectors. More details are provided in Algorithm~\ref{alg:param_gen}.
\section{Experiments}
\label{sec:experiments}

\subsection{Experimental Setup}

\paragraph{Implementation Details.}
All experiments are performed on a NVIDIA RTX~V100 GPU. We train \ourmethod\ with AdamW and a linear learning rate schedule, starting from $1\times10^{-4}$. 

\paragraph{Datasets and Model Collection.}
We evaluate \ourmethod\ on a broad suite of pretrained models gathered from public repositories, covering diverse architectures and datasets.  This setup enables a thorough examination of \ourmethod's performance across different model scales, and data settings. The instructions used to guide the  weights generation consist of text description of the architecture combined with dataset embeddings.

\textbf{Evaluation Protocol}
We evaluate \ourmethod\ through three primary experiments:
\begin{enumerate}
    \item Comparison with existing methods on the pretrained model zoo from~\citet{Schrholt2022HyperRepresentationsAG}
    \item Generalization assessment across diverse ResNet architectures
    \item Parameter generation efficiency evaluation on architectures with varying parameter counts
\end{enumerate}

\textbf{Baselines}
We compare IGPG against three state-of-the-art approaches:
\begin{wraptable}[29]{r}{0.6\textwidth}
\caption{Comparison of weight initialization methods trained on  pretrained from epochs 21--25. We compare: (1) training from scratch (tr. fr. scratch), (2) $S_{\mathrm{KDE30}}$~\citep{schrholt2022hyperrepresentation}, (3) SANE with $KDE30$, (4) subsampled SANE$_{\mathrm{SUB}}$ (aligned with \ourmethod), and (5) D2NWG~\citep{soro2024diffusionbased}.}

\label{tab:hyperzoo}
\small
\setlength{\tabcolsep}{1.0pt}
\begin{tabular}{clcccc}
\toprule
Ep.                  & \multicolumn{1}{c}{Method} & MNIST               & SVHN               & CIFAR-10           & STL                \\
\cmidrule(r){1-1} \cmidrule(lr){2-2} \cmidrule(lr){3-3} \cmidrule(lr){4-4} \cmidrule(lr){5-5} \cmidrule(l){6-6}
\multirow{5}{*}{0}     &  tr. fr. scratch                  & $\sim$10 /\%        & $\sim$10 /\%       & $\sim$10 /\%       & $\sim$10 /\%       \\
                       & $S_{KDE30}$                & 68.6$\pm$6.7           & 54.5$\pm$5.9          & \textit{n/a}                & \textit{n/a}                \\
                       & $SANE_{KDE30}$                    & 84.8$\pm$0.8           & 70.7$\pm$1.4          & 56.3$\pm$0.5          & 39.2$\pm$0.8          \\
                       & $SANE_{SUB}$                   & \textbf{86.7$\pm$0.8}  & \textbf{72.3$\pm$1.6} & 57.9$\pm$0.2 & 43.5$\pm$1.0 \\


 & D2NWG                & 80.52$\pm$0.82                & 66.6$\pm$0.7    & 58.80$\pm$0.1  & 44.50$\pm$0.1              \\
                       & \ourmethod                 & 83.20$\pm$0.01                & 67.10$\pm$0.4              & \textbf{58.3$\pm$0.1}               & \textbf{44.41$\pm$0.1}              \\
\cmidrule(r){1-1} \cmidrule(lr){2-2} \cmidrule(lr){3-3} \cmidrule(lr){4-4} \cmidrule(lr){5-5} \cmidrule(l){6-6}
\multirow{5}{*}{1}     &  tr. fr. scratch                  & 20.6$\pm$1.6           & 19.4$\pm$0.6          & 37.2$\pm$1.4          & 21.3$\pm$1.6          \\
                       & $S_{KDE30}$                & 83.7$\pm$1.3           & 69.9$\pm$1.6          & \textit{n/a}                & \textit{n/a}                \\
                       & $SANE_{KDE30}$                    & 85.5$\pm$0.8           & 71.3$\pm$1.4          & 58.2$\pm$0.2          & 43.5$\pm$0.7          \\
                       & $SANE_{SUB}$                   & \textbf{87.5$\pm$0.6}  & 73.3$\pm$1.4 & \textbf{59.1$\pm$0.3} & 44.3$\pm$1.0 \\


          & D2NWG              & 87.8$\pm$0.4 & 73.6$\pm$1.3  & 59.2$\pm$0.3       & 44.8$\pm$0.2               \\
                                  & \ourmethod                 & 86.10$\pm$0.1               & \textbf{74.0$\pm$0.3}                & 58.7$\pm$0.3               & \textbf{44.9$\pm$0.1}              \\
\cmidrule(r){1-1} \cmidrule(lr){2-2} \cmidrule(lr){3-3} \cmidrule(lr){4-4} \cmidrule(lr){5-5} \cmidrule(l){6-6}
\multirow{5}{*}{5}     &  tr. fr. scratch                  & 36.7$\pm$5.2           & 23.5$\pm$4.7          & 48.5$\pm$1.0          & 31.6$\pm$4.2          \\
                       & $S_{KDE30}$                & 92.4$\pm$0.7 & 57.3$\pm$12.4         & \textit{n/a}                & \textit{n/a}                \\
                       & $SANE_{KDE30}$                    & 87.5$\pm$0.7           & 72.2$\pm$1.2          & 58.8$\pm$0.4          & 45.2$\pm$0.6          \\
                       & $SANE_{SUB}$                   & 89.0$\pm$0.4           & 73.6$\pm$1.5 & 59.6$\pm$0.3 & 45.3$\pm$0.9 \\

                &  D2NWG                    &  \textbf{92.5$\pm$0.9}  & 74.0$\pm$0.1  & 60.3$\pm$0.1      & 45.4$\pm$0.1               \\
                                       & \ourmethod                 & 91.7$\pm$0.8                & \textbf{74.5$\pm$0.5}               & \textbf{60.3$\pm$0.1}               & \textbf{45.7$\pm$0.1}               \\
\cmidrule(r){1-1} \cmidrule(lr){2-2} \cmidrule(lr){3-3} \cmidrule(lr){4-4} \cmidrule(lr){5-5} \cmidrule(l){6-6}

\multirow{5}{*}{25}    &  tr. fr. scratch                  & 83.3$\pm$2.6           & 66.7$\pm$8.5          & 57.2$\pm$0.8          & 44.0$\pm$1.0          \\
                       & $S_{KDE30}$                & 93.0$\pm$0.7           & 74.2$\pm$1.4          &       \textit{n/a}             &    \textit{n/a}                \\
                       & $SANE_{KDE30}$                    & 92.0$\pm$0.3           & 74.7$\pm$0.8          & 60.2$\pm$0.6          & 48.4$\pm$0.5 \\
                       & $SANE_{SUB}$                   & 92.3$\pm$0.4           & 75.1$\pm$1.0 & 61.2$\pm$0.1 & 48.0$\pm$0.4          \\

                & D2NWG                 &  \textbf{96.2$\pm$0.3}      &  75.7$\pm$0.5  & \textbf{64.1$\pm$1.0}    & 48.7$\pm$0.5        \\
                               & \ourmethod                 & 94.5$\pm$0.1       & \textbf{76.9$\pm$0.1 }              & 63.9$\pm$0.0               & \textbf{49.06$\pm$0.2}              \\
\cmidrule(r){1-1} \cmidrule(lr){2-2} \cmidrule(lr){3-3} \cmidrule(lr){4-4} \cmidrule(lr){5-5} \cmidrule(l){6-6}
\multicolumn{1}{l}{50} &  tr. fr. scratch                  & 91.1$\pm$2.6           & 70.7$\pm$8.8          & 61.5$\pm$0.7          & 47.4$\pm$0.9      \\
\bottomrule
\end{tabular}
\end{wraptable} 

\begin{itemize}
    \item Hyper-representations~\citep{Schrholt2022HyperRepresentationsAG}$(S_{KDE}$: A weights generation method that uses kernel density estimator(KDE) as prior.
    \item SANE~\citep{schuerholt2024sane}: An improved version of $(S_{KDE}$ that uses weight tokenization with a KDE prior.
    \item D2NWG~\citep{soro2024diffusionbased}: A diffusion-based approach to neural network weight generation.
\end{itemize}

\subsection{Benchmarking on Tiny Model Zoo}\label{zoo}

We evaluate \ourmethod\ on the Tiny Model Zoo dataset~\citep{schurholt2022model}, which comprises compact CNNs trained on MNIST, SVHN, CIFAR-10, and STL-10. Specifically, we use a 2-layer CNN (2{,}464 parameters) for MNIST and SVHN, and a 3-layer CNN (10{,}853 parameters) for CIFAR-10 and STL-10. Following prior work, we draw pretrained weights from epochs 21--25 of a 50-epoch training schedule, with datasets split into training (70\%), validation (15\%), and test (15\%). Unlike methods requiring separate models per dataset~\citep{Schrholt2022HyperRepresentationsAG,schuerholt2024sane}, \ourmethod\ learns a single generator that robustly handles all architectures and tasks.
\paragraph{Task.}
We evaluate IGPG’s ability to generate neural network parameters that remain effective under both fine-tuning and transfer learning scenarios. In particular, we seek to confirm that IGPG’s synthesized weights are readily adaptable for fine-tuning scenarios.

\paragraph{Results and Analysis.}
Table~\ref{tab:hyperzoo} shows that IGPG outperforms the sequential parameter generation method of \citet{schuerholt2024sane}, while matching the rapid convergence characteristic of state-of-the-art diffusion-based approaches. Crucially, IGPG preserves both zero-shot accuracy and fine-tuning performance when compared to previous works~\citep{schuerholt2024sane,Schrholt2022HyperRepresentationsAG}. This highlights IGPG’s capacity to generate robust initial weights that can be efficiently adapted, thereby accommodating multiple architectures and datasets within a single unified framework.

\subsection{Transfer Learning and Fine-tuning on Unseen Datasets}
In this section, we train a single model on 30 diverse datasets from Meta-Album~\citep{meta-album-2022}, which span a broad range of class distributions. We then sample parameters for CIFAR-10 and Oxford Pets, thus evaluating how well a model pretrained on heterogeneous datasets adapts to unseen tasks. The training and target datasets are disjoint to ensure a fair assessment.

\begin{figure}[t]

        \centering
        \includegraphics[width=1.0\linewidth]{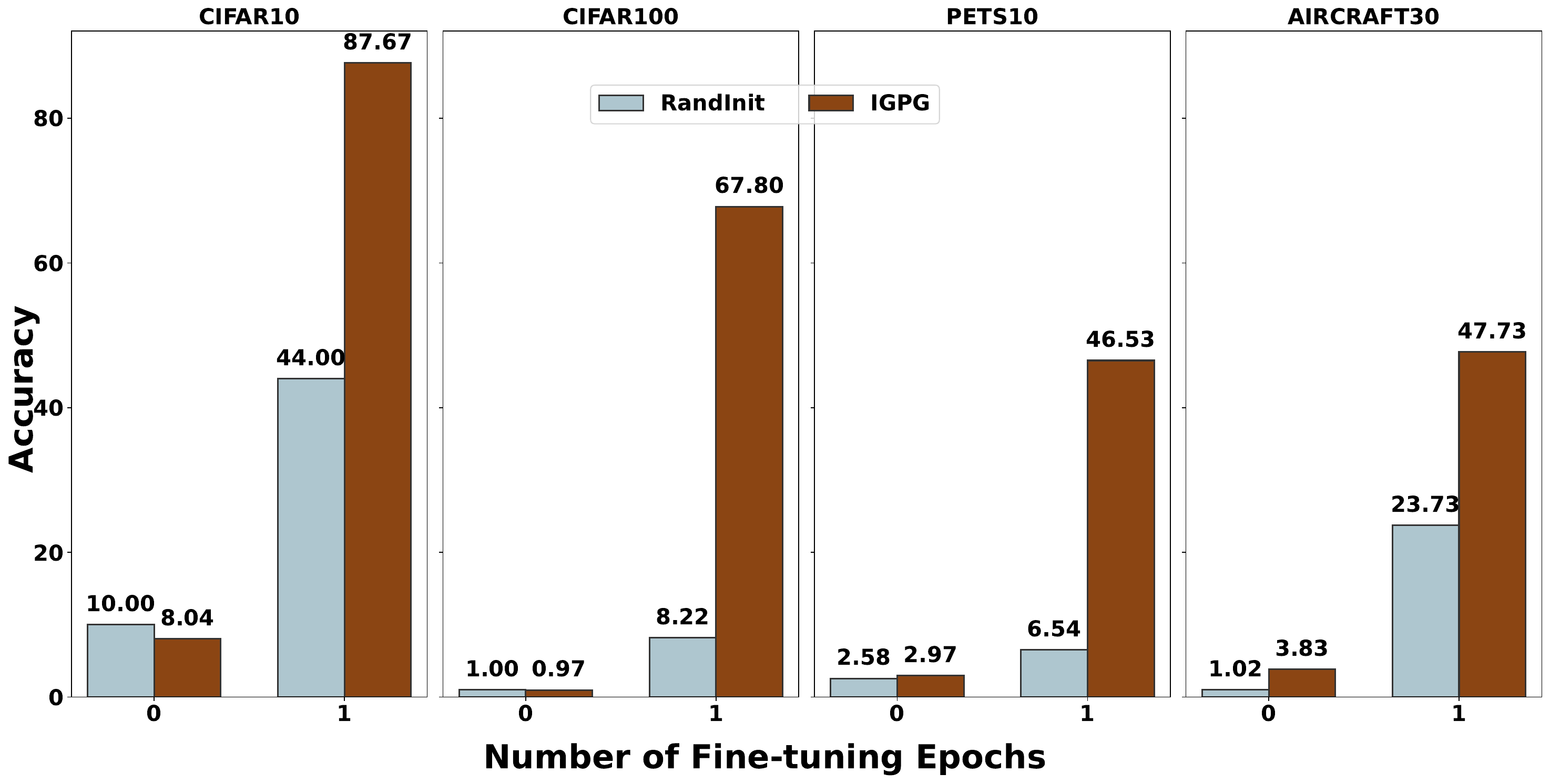}
        \caption{\small Transfer learning evaluation on novel datasets: CIFAR100, CIFAR10, Aircraft30, and PETS10 compared to random initialization.}
        \label{fig:transfer_learning}

\end{figure}

Because we fix the architecture (a MobileNetV3 subnet from OFA) and pretrained with images of size $224$, we only encode dataset information rather than architectural details since the arhitecture is the same. Table~\ref{tab:datasets} lists the training datasets. As shown in Figure~\ref{fig:transfer_learning}, our sampled models begin at a performance level on par with random initialization but achieve over 50\% relative improvement within one epoch. This underscores the advantage of leveraging broad pretraining data for faster adaptation: although zero-shot performance may start near baseline, it quickly surpasses random initialization, reflecting the effectiveness of our method in generating meaningful parameters.

\subsection{Cross-Architecture Benchmarking}

We evaluated our instruction-guided parameter generation on CIFAR-10 using 125 randomly sampled ResNet-56 variants spanning 200k--700k parameters, with block configurations from [4,4,4] to [8,8,8]. Each model was trained for 100~epochs, and we collected the last five epochs' weights to form our training set. We set the maximum token-sequence length to 768 and trained a VQ-VAE to encode these parameters. Next, we employed a GPT-2 model, conditioned by an instruction template (preprocessed via LLaMA3-1-8B-Instruct), to generate the codebook tokens. We also experimented with fine-tuning larger language models on these codebooks, observing that while minor token mismatches (e.g.\ non-integers) can occur, the approach remains feasible.

\begin{figure}[t]
\centering
\includegraphics[width=0.8\linewidth]{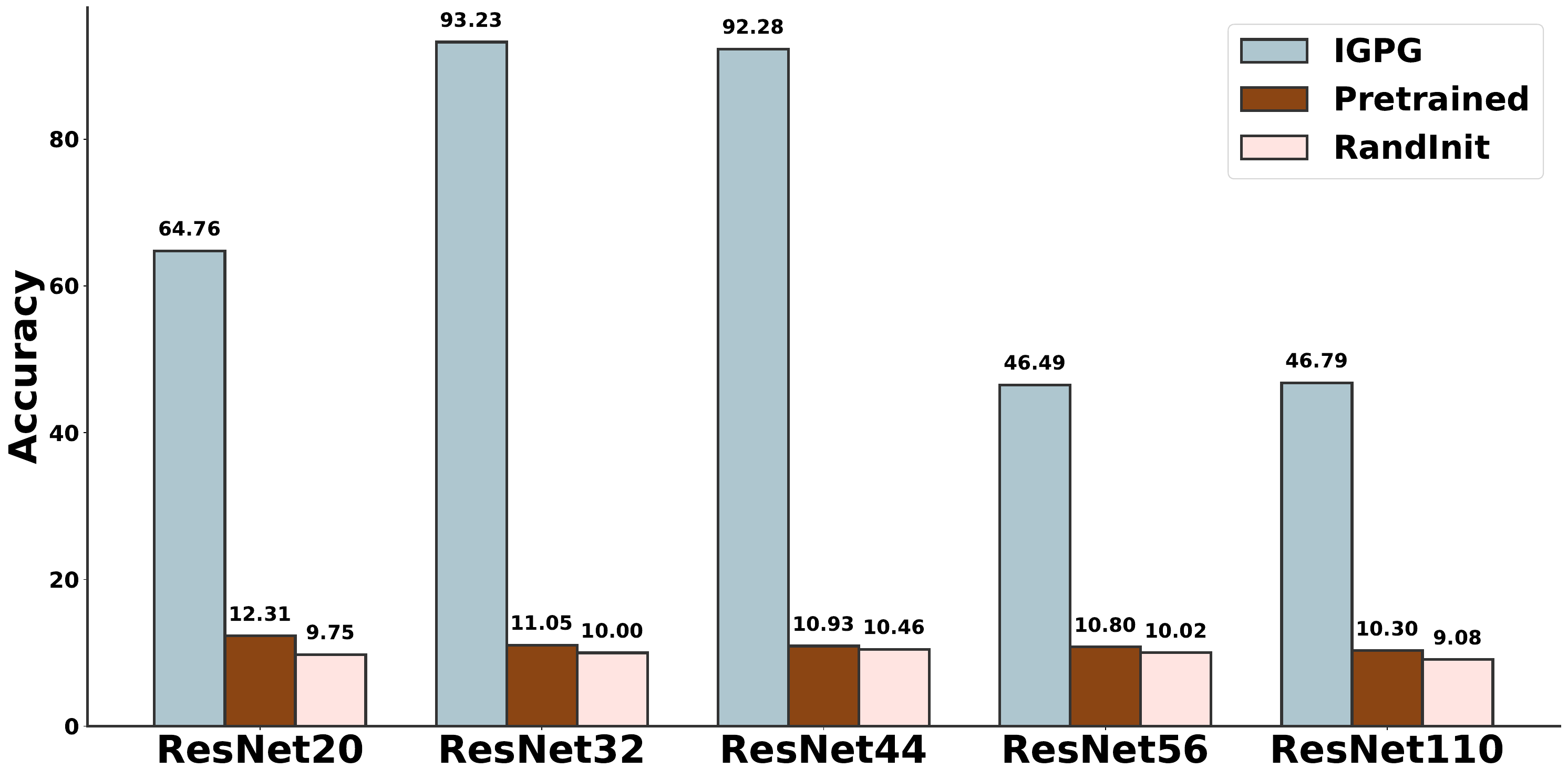}
    \caption[]{\small Performance evaluation with seen and unseen ResNet architectures on CIFAR-10 against models pretrained on CIFAR-100 and Random Initialization.}
    \vspace{-0.15in}
    \label{bar_cross_res}
\end{figure}

We then tested five ResNet architectures, including two in-distribution variants (directly sampled) and three out-of-distribution (ResNet-20, ResNet-56, and ResNet-110). Figure~\ref{bar_cross_res} compares \ourmethod\ against random initialization and CIFAR-100-pretrained weights. On in-distribution architectures, \ourmethod\ achieves accuracy on par with pretrained baselines. For out-of-distribution networks, our method attains up to 64\% accuracy on ResNet-20 and 46\% on both ResNet-56 and ResNet-110, outperforming the other baselines. These results highlight \ourmethod's strong cross-architecture generalization without requiring additional fine-tuning, underscoring the potential of instruction-guided parameter generation in handling unseen network configurations.

\begin{figure}[t!]
    \centering
     \begin{subfigure}{0.4\textwidth}

            \includegraphics[width=1.0\textwidth]{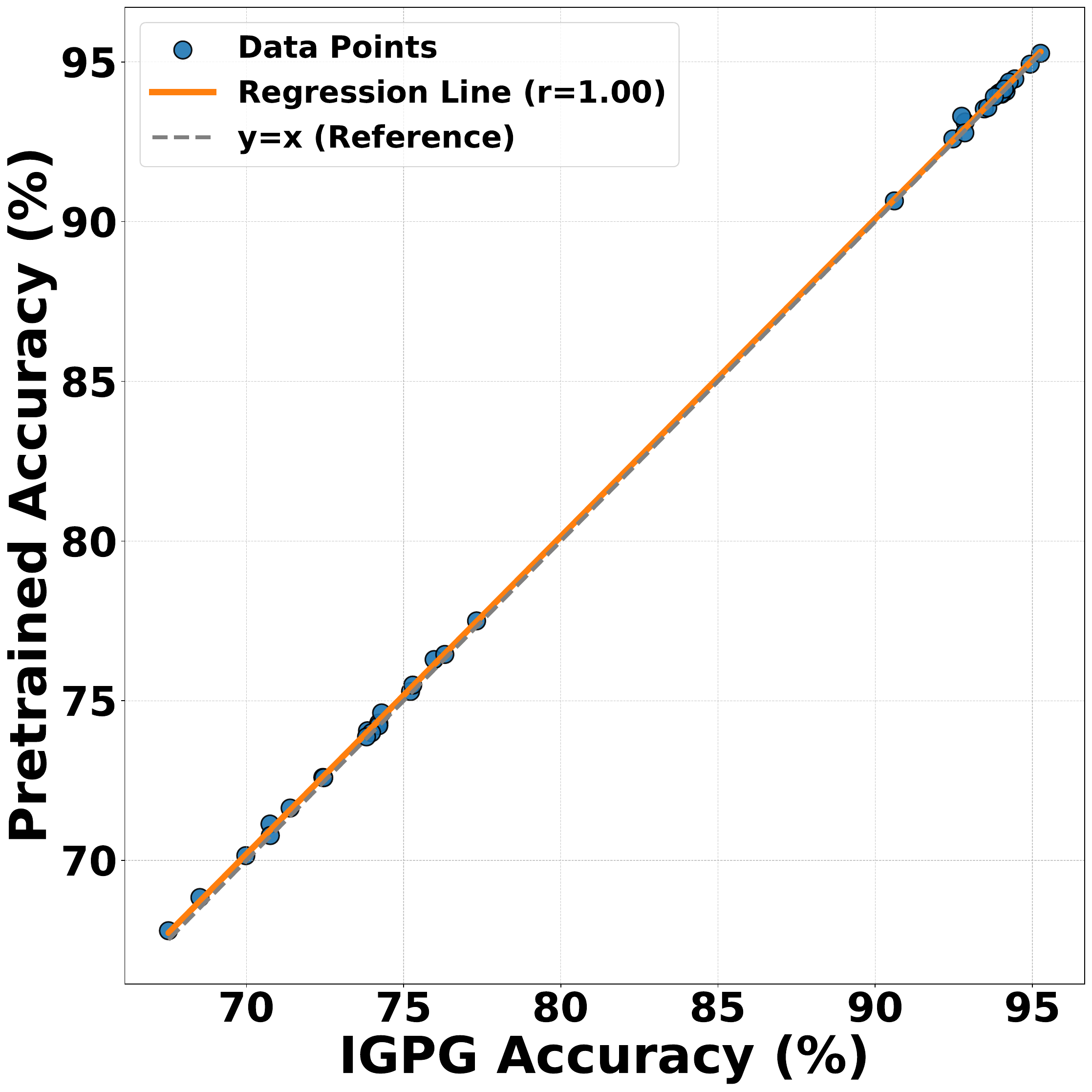}
            \caption{CIFAR10}
            \label{figmnist}
    \end{subfigure}
     \begin{subfigure}{0.4\textwidth}

            \includegraphics[width=1.0\textwidth]{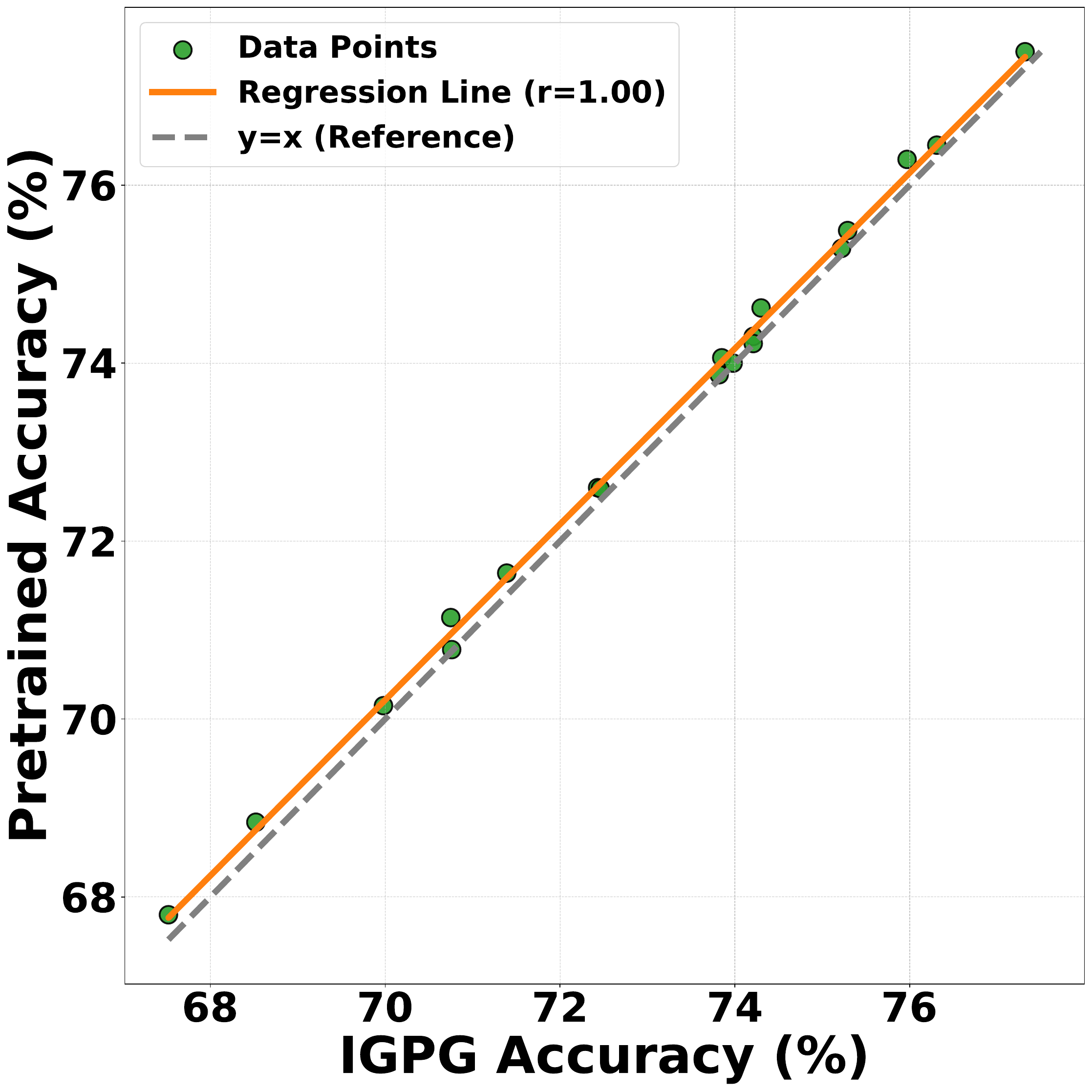}
            \caption{CIFAR100}
            \label{figsvhn}
    \end{subfigure}
        \caption{Comparison of \ourmethod's conditional sampled weight based initialization versus pretrained models across diverse architectures on CIFAR10 and CIFAR100}
        \label{fig:corr}
	\vspace{-0.15in}
\end{figure}

\subsection{Handling Diverse Pretrained Models from Varied Datasets}
\label{multiples}

We further demonstrate \ourmethod's versatility by encoding a broad set of architectures pretrained on datasets with varying numbers of classes (CIFAR-10 vs.\ CIFAR-100). We gather 19 publicly available models from GitHub\footnote{\url{https://github.com/chenyaofo/pytorch-cifar-models}} spanning ResNet, ShuffleNet, MobileNet, and others. These architectures range from 0.27M to 27M parameters (over 100$\times$ difference), as shown in Figure~\ref{bar_param}.
\paragraph{Experimental Setup.}
We train a VQ-VAE with a chunk size of 2694256 parameters, each encoded into 64 tokens. For the largest models (roughly 10 chunks), this translates into a maximum sequence length of 640 tokens for our autoregressive model. We condition on both architecture descriptions and dataset encodings (via a CLIP image encoder using five images per class).

\paragraph{Results and Significance.}
Figure~\ref{fig:corr} reports near-perfect Pearson correlations for both CIFAR-10 (0.9999) and CIFAR-100 (0.9991), suggesting that our generated parameters track closely with original pretrained weights. The regression lines for each dataset align closely with $y=x$, indicating comparable performance between \ourmethod-generated weights and their pretrained counterparts. These findings highlight \ourmethod's capacity to learn from diverse architecture--dataset pairs and produce parameters that faithfully approximate original performance, pointing to its potential for guiding model selection and fast adaptation under dataset- or instruction-based constraints.

\subsection{Extension to Diverse LoRA Parameters Generation}
To demonstrate that \ourmethod{} can learn distributions of diverse LoRA modules and improve downstream performance, we evaluated it on six standard computer vision benchmarks: Oxford-IIIT Pets, Stanford Cars, CIFAR-10, EuroSAT, the Describable Textures Dataset (DTD), and FGVC Aircraft. We began by fine-tuning a Vision Transformer (ViT-Base) with LoRA modules, following the procedure of \citet{gao2024parameterefficientfinetuningdiscretefourier}. For each dataset, we retained the top five performing models and learned their parameter distributions. We then used a specialized version of \ourmethod{} to explore these distributions and generate novel LoRA parameters.

Table~\ref{tab:cv} shows that our generated LoRA parameters deliver up to a 10\% improvement compared to the baseline pretrained model~\citep{gao2024parameterefficientfinetuningdiscretefourier}, highlighting the efficacy of our distribution-learning approach in uncovering higher-performing LoRA configurations within the existing parameter space.

\begin{table}[t]
\centering
\caption{ Performance evaluation on LoRA weights generation conditioned on the dataset}
\label{tab:cv}
\addtolength{\tabcolsep}{-4.3pt}
\resizebox{1.0\textwidth}{!}{%
\begin{tabular}{@{}c|l|r|ccccccc@{}}
\toprule
Model & Method & \begin{tabular}[c]{@{}r@{}}\# Trainable\\ Parameters\end{tabular} & OxfordPets & StanfordCars & CIFAR10 & DTD & EuroSAT & FGVC  & Average \\
\midrule
\multirow{5}{*}{\rotatebox{90}{ViT-Base}} & LP & - & 90.28\textsubscript{$\pm$0.43} & 25.76\textsubscript{$\pm$0.28} & 96.41\textsubscript{$\pm$0.02} & 69.77\textsubscript{$\pm$0.67} & 88.72\textsubscript{$\pm$0.13} & 17.44\textsubscript{$\pm$0.43} & 64.73\\

 & FF & 85.8M & 93.14\textsubscript{$\pm$0.40} & \textbf{79.78}\textsubscript{$\pm$1.15} & \textbf{98.92}\textsubscript{$\pm$0.05} & 77.68\textsubscript{$\pm$1.21} & \textbf{99.05}\textsubscript{$\pm$0.09} & \textbf{54.84}\textsubscript{$\pm$1.23} &  \textbf{83.90}\\
 
 & LoRA & 581K & 93.19\textsubscript{$\pm$0.36} & 45.38\textsubscript{$\pm$0.41} & 98.78\textsubscript{$\pm$0.05} & 74.95\textsubscript{$\pm$0.40} & 98.44\textsubscript{$\pm$0.15} & 25.16\textsubscript{$\pm$0.16}   & 72.65\\
 
 & FourierFT\citep{gao2024parameterefficientfinetuningdiscretefourier} & 72K & \textbf{93.21}\textsubscript{$\pm$0.26} & 46.11\textsubscript{$\pm$0.24} & 98.58\textsubscript{$\pm$0.07} & 75.09\textsubscript{$\pm$0.37} & 98.29\textsubscript{$\pm$0.04} & 27.51\textsubscript{$\pm$0.64}  &  73.13\\

 \midrule
 & \textbf{\ourmethod} & 72K & 92.84\textsubscript{$\pm$0.30} & 57.67\textsubscript{$\pm$0.52} & 98.45\textsubscript{$\pm$0.95} & \textbf{88.74}\textsubscript{$\pm$0.08} & 98.70\textsubscript{$\pm$0.23} & 36.63\textsubscript{$\pm$0.11} &  78.83\\ 

\bottomrule
\end{tabular}%
}
\end{table}
\subsection{Learning Distribution of models pretrained on Diverse Datasets}
To demonstrate that IGPG  can learn distribution of multiple model pretrained on both large, medium and small scale dataset while maintaining a higher performance, we collect pretrained weights of ViT and Mobilenetv3 small pretrained on various datasets including ImageNet-1k(see Table ~\ref{tab:weight-generation-results}) to train our method. 
\paragraph{Experiment Setup and Findings.} 
We train VQ-VAE on combined weights from diverse datasets, and the transformer on its codebook using 5 samples per class and architecture descriptions as instructions, then evaluate both reconstruction and autoregressive sampling in-distribution.  Table~\ref{tab:weight-generation-results} reports the results as follows:
\textit{Pretrained (\%)}: Accuracy of the original pretrained weights. \textit{VQ-VAE Reconstruction (\%)}: Accuracy after encoding and decoding the pretrained weights via VQ-VAE, showing the fidelity of our compressive representation. \textit{Best Sampled Weights (\%) with \ourmethod}: Accuracy of the best sample among five randomly drawn sequences from our trained model, conditioned on the respective dataset and architecture. 
For ViT-Small across CIFAR-10, CIFAR-100, CINIC-10, SVHN, and Tiny-ImageNet, we observe that both VQ-VAE reconstruction and IGPG sampled weights attain accuracy nearly indistinguishable from the original pretrained models. Moreover, the best samples often match or slightly exceed the baseline. On MobileNetV3-Small, sampled weights remain competitive with the pretrained baseline. These results confirm that IGPG  preserves performance while offering the flexibility to sample diverse parameter configurations.

\begin{table}[t]
\caption{Learning distribution of combined ViT-Small (CIFAR-10, CIFAR-100, CINIC-10, SVHN, Tiny-ImageNet) and MobileNetV3-Small (ImageNet). We report model size (Parameters), Pretrained accuracy, VQVAE Reconstruction accuracy, and the best among five samples (Best Sampled Weights).  }
\label{tab:weight-generation-results}
\centering
\resizebox{\columnwidth}{!}{%
\begin{tabular}{llcccc}
\toprule
Dataset       & Architecture   & Parameters & Pretrained (\%) & VQVAE Reconstruction (\%) & \ourmethod (\%) \\ 
\midrule
CIFAR-10          & ViT-Small               & 2697610      & 93.67             & 93.67                        & \textbf{93.70}                       \\ 
CIFAR-100         & ViT-Small               & 2714980      & 72.29             & 72.28                        & \textbf{72.30}                       \\ 
CINIC-10          & ViT-Small               & 2697610      & 83.23             & 83.23                        & \textbf{83.24}                       \\ 
SVHN              & ViT-Small               & 2697610      & 97.74             & 97.74                        & \textbf{97.76}                       \\ 
Tiny-ImageNet     & ViT-Small               & 2771144      & 53.44    & 53.44                        & \textbf{53.45}                       \\ 
ImageNet          & MobileNetV3-Small       & 2542856      & \textbf{67.67}    & 67.66                        & 67.65                       \\ 
\bottomrule
\end{tabular}%
}
\end{table}
Additional results are presented in the Appendix~\ref{add} and Ablation in Appendix~\ref{ablat}.

\section{Conclusion}
We introduced IGPG, an instruction-guided framework for neural network parameter generation that combines a VQVAE with autoregressive modeling. IGPG generates network parameters conditioned on task descriptions and architectural specifications. Experimental results demonstrate that IGPG achieves performance comparable to that of pretrained models while converging faster than random initialization. Furthermore, our approach effectively compresses large pretrained datasets and generalizes across diverse architectures, thereby advancing the fields of neural architecture search and transfer learning.

\subsubsection*{Acknowledgments}
This work was supported by Institute for Information \& communications Technology Planning \& Evaluation(IITP) grant funded by the Korea government(MSIT) (RS-2019-II190075, Artificial Intelligence Graduate School Program(KAIST)) and (No.RS-2022-II220713, Meta-learning Applicable to Real-world Problems), by Samsung Research Funding Center of Samsung Electronics (No. IO201210-08006-01), Institute of Information \& communications Technology Planning \& Evaluation (IITP) under Open RAN Education and Training Program (IITP-2024-RS-2024-00429088) grant funded by the Korea government(MSIT) and, by National Research Foundation of Korea (NRF) grant funded by the Korea government (MSIT) (No. RS-2023-00256259)

\bibliography{./iclr2025_conference}

\begin{thebibliography}{32}
\providecommand{\natexlab}[1]{#1}
\providecommand{\url}[1]{\texttt{#1}}
\expandafter\ifx\csname urlstyle\endcsname\relax
  \providecommand{\doi}[1]{doi: #1}\else
  \providecommand{\doi}{doi: \begingroup \urlstyle{rm}\Url}\fi

\bibitem[Addair \& Rishi(2024)Addair and Rishi]{loraland2024}
Justin Zhao Timothy Wang Wael Abid Geoffrey Angus Arnav Garg Jeffery Kinnison Piero Molino~Travis Addair and Devvret Rishi.
\newblock Lora land: 310 fine-tuned llms that rival gpt-4, a technical report, April 2024.
\newblock URL \url{https://predibase.com/blog/lora-land-fine-tuned-open-source-llms-that-outperform-gpt-4}.

\bibitem[Deutsch(2018)]{Deutsch2018GeneratingNN}
Lior Deutsch.
\newblock Generating neural networks with neural networks.
\newblock \emph{ArXiv}, abs/1801.01952, 2018.

\bibitem[Dubey et~al.(2024)Dubey, Jauhri, Pandey, Kadian, Al-Dahle, Letman, Mathur, Schelten, Yang, Fan, Goyal, Hartshorn, Yang, Mitra, Sravankumar, Korenev, Hinsvark, Rao, Zhang, Rodriguez, Gregerson, Spataru, Roziere, Biron, Tang, Chern, Caucheteux, Nayak, Bi, Marra, McConnell, Keller, Touret, Wu, Wong, Ferrer, Nikolaidis, Allonsius, Song, Pintz, Livshits, Esiobu, Choudhary, Mahajan, Garcia-Olano, Perino, Hupkes, Lakomkin, AlBadawy, Lobanova, Dinan, Smith, Radenovic, Zhang, Synnaeve, Lee, Anderson, Nail, Mialon, Pang, Cucurell, Nguyen, Korevaar, Xu, Touvron, Zarov, Ibarra, Kloumann, Misra, Evtimov, Copet, Lee, Geffert, Vranes, Park, Mahadeokar, Shah, van~der Linde, Billock, Hong, Lee, Fu, Chi, Huang, Liu, Wang, Yu, Bitton, Spisak, Park, Rocca, Johnstun, Saxe, Jia, Alwala, Upasani, Plawiak, Li, Heafield, Stone, El-Arini, Iyer, Malik, Chiu, Bhalla, Rantala-Yeary, van~der Maaten, Chen, Tan, Jenkins, Martin, Madaan, Malo, Blecher, Landzaat, de~Oliveira, Muzzi, Pasupuleti, Singh, Paluri, Kardas, Oldham, Rita,
  Pavlova, Kambadur, Lewis, Si, Singh, Hassan, Goyal, Torabi, Bashlykov, Bogoychev, Chatterji, Duchenne, Çelebi, Alrassy, Zhang, Li, Vasic, Weng, Bhargava, Dubal, Krishnan, Koura, Xu, He, Dong, Srinivasan, Ganapathy, Calderer, Cabral, Stojnic, Raileanu, Girdhar, Patel, Sauvestre, Polidoro, Sumbaly, Taylor, Silva, Hou, Wang, Hosseini, Chennabasappa, Singh, Bell, Kim, Edunov, Nie, Narang, Raparthy, Shen, Wan, Bhosale, Zhang, Vandenhende, Batra, Whitman, Sootla, Collot, Gururangan, Borodinsky, Herman, Fowler, Sheasha, Georgiou, Scialom, Speckbacher, Mihaylov, Xiao, Karn, Goswami, Gupta, Ramanathan, Kerkez, Gonguet, Do, Vogeti, Petrovic, Chu, Xiong, Fu, Meers, Martinet, Wang, Tan, Xie, Jia, Wang, Goldschlag, Gaur, Babaei, Wen, Song, Zhang, Li, Mao, Coudert, Yan, Chen, Papakipos, Singh, Grattafiori, Jain, Kelsey, Shajnfeld, Gangidi, Victoria, Goldstand, Menon, Sharma, Boesenberg, Vaughan, Baevski, Feinstein, Kallet, Sangani, Yunus, Lupu, Alvarado, Caples, Gu, Ho, Poulton, Ryan, Ramchandani, Franco, Saraf,
  Chowdhury, Gabriel, Bharambe, Eisenman, Yazdan, James, Maurer, Leonhardi, Huang, Loyd, Paola, Paranjape, Liu, Wu, Ni, Hancock, Wasti, Spence, Stojkovic, Gamido, Montalvo, Parker, Burton, Mejia, Wang, Kim, Zhou, Hu, Chu, Cai, Tindal, Feichtenhofer, Civin, Beaty, Kreymer, Li, Wyatt, Adkins, Xu, Testuggine, David, Parikh, Liskovich, Foss, Wang, Le, Holland, Dowling, Jamil, Montgomery, Presani, Hahn, Wood, Brinkman, Arcaute, Dunbar, Smothers, Sun, Kreuk, Tian, Ozgenel, Caggioni, Guzmán, Kanayet, Seide, Florez, Schwarz, Badeer, Swee, Halpern, Thattai, Herman, Sizov, Guangyi, Zhang, Lakshminarayanan, Shojanazeri, Zou, Wang, Zha, Habeeb, Rudolph, Suk, Aspegren, Goldman, Damlaj, Molybog, Tufanov, Veliche, Gat, Weissman, Geboski, Kohli, Asher, Gaya, Marcus, Tang, Chan, Zhen, Reizenstein, Teboul, Zhong, Jin, Yang, Cummings, Carvill, Shepard, McPhie, Torres, Ginsburg, Wang, Wu, U, Saxena, Prasad, Khandelwal, Zand, Matosich, Veeraraghavan, Michelena, Li, Huang, Chawla, Lakhotia, Huang, Chen, Garg, A, Silva, Bell,
  Zhang, Guo, Yu, Moshkovich, Wehrstedt, Khabsa, Avalani, Bhatt, Tsimpoukelli, Mankus, Hasson, Lennie, Reso, Groshev, Naumov, Lathi, Keneally, Seltzer, Valko, Restrepo, Patel, Vyatskov, Samvelyan, Clark, Macey, Wang, Hermoso, Metanat, Rastegari, Bansal, Santhanam, Parks, White, Bawa, Singhal, Egebo, Usunier, Laptev, Dong, Zhang, Cheng, Chernoguz, Hart, Salpekar, Kalinli, Kent, Parekh, Saab, Balaji, Rittner, Bontrager, Roux, Dollar, Zvyagina, Ratanchandani, Yuvraj, Liang, Alao, Rodriguez, Ayub, Murthy, Nayani, Mitra, Li, Hogan, Battey, Wang, Maheswari, Howes, Rinott, Bondu, Datta, Chugh, Hunt, Dhillon, Sidorov, Pan, Verma, Yamamoto, Ramaswamy, Lindsay, Lindsay, Feng, Lin, Zha, Shankar, Zhang, Zhang, Wang, Agarwal, Sajuyigbe, Chintala, Max, Chen, Kehoe, Satterfield, Govindaprasad, Gupta, Cho, Virk, Subramanian, Choudhury, Goldman, Remez, Glaser, Best, Kohler, Robinson, Li, Zhang, Matthews, Chou, Shaked, Vontimitta, Ajayi, Montanez, Mohan, Kumar, Mangla, Albiero, Ionescu, Poenaru, Mihailescu, Ivanov, Li, Wang,
  Jiang, Bouaziz, Constable, Tang, Wang, Wu, Wang, Xia, Wu, Gao, Chen, Hu, Jia, Qi, Li, Zhang, Zhang, Adi, Nam, Yu, Wang, Hao, Qian, He, Rait, DeVito, Rosnbrick, Wen, Yang, and Zhao]{dubey2024llama3herdmodels}
Abhimanyu Dubey, Abhinav Jauhri, Abhinav Pandey, Abhishek Kadian, Ahmad Al-Dahle, Aiesha Letman, Akhil Mathur, Alan Schelten, Amy Yang, Angela Fan, Anirudh Goyal, Anthony Hartshorn, Aobo Yang, Archi Mitra, Archie Sravankumar, Artem Korenev, Arthur Hinsvark, Arun Rao, Aston Zhang, Aurelien Rodriguez, Austen Gregerson, Ava Spataru, Baptiste Roziere, Bethany Biron, Binh Tang, Bobbie Chern, Charlotte Caucheteux, Chaya Nayak, Chloe Bi, Chris Marra, Chris McConnell, Christian Keller, Christophe Touret, Chunyang Wu, Corinne Wong, Cristian~Canton Ferrer, Cyrus Nikolaidis, Damien Allonsius, Daniel Song, Danielle Pintz, Danny Livshits, David Esiobu, Dhruv Choudhary, Dhruv Mahajan, Diego Garcia-Olano, Diego Perino, Dieuwke Hupkes, Egor Lakomkin, Ehab AlBadawy, Elina Lobanova, Emily Dinan, Eric~Michael Smith, Filip Radenovic, Frank Zhang, Gabriel Synnaeve, Gabrielle Lee, Georgia~Lewis Anderson, Graeme Nail, Gregoire Mialon, Guan Pang, Guillem Cucurell, Hailey Nguyen, Hannah Korevaar, Hu~Xu, Hugo Touvron, Iliyan Zarov,
  Imanol~Arrieta Ibarra, Isabel Kloumann, Ishan Misra, Ivan Evtimov, Jade Copet, Jaewon Lee, Jan Geffert, Jana Vranes, Jason Park, Jay Mahadeokar, Jeet Shah, Jelmer van~der Linde, Jennifer Billock, Jenny Hong, Jenya Lee, Jeremy Fu, Jianfeng Chi, Jianyu Huang, Jiawen Liu, Jie Wang, Jiecao Yu, Joanna Bitton, Joe Spisak, Jongsoo Park, Joseph Rocca, Joshua Johnstun, Joshua Saxe, Junteng Jia, Kalyan~Vasuden Alwala, Kartikeya Upasani, Kate Plawiak, Ke~Li, Kenneth Heafield, Kevin Stone, Khalid El-Arini, Krithika Iyer, Kshitiz Malik, Kuenley Chiu, Kunal Bhalla, Lauren Rantala-Yeary, Laurens van~der Maaten, Lawrence Chen, Liang Tan, Liz Jenkins, Louis Martin, Lovish Madaan, Lubo Malo, Lukas Blecher, Lukas Landzaat, Luke de~Oliveira, Madeline Muzzi, Mahesh Pasupuleti, Mannat Singh, Manohar Paluri, Marcin Kardas, Mathew Oldham, Mathieu Rita, Maya Pavlova, Melanie Kambadur, Mike Lewis, Min Si, Mitesh~Kumar Singh, Mona Hassan, Naman Goyal, Narjes Torabi, Nikolay Bashlykov, Nikolay Bogoychev, Niladri Chatterji, Olivier
  Duchenne, Onur Çelebi, Patrick Alrassy, Pengchuan Zhang, Pengwei Li, Petar Vasic, Peter Weng, Prajjwal Bhargava, Pratik Dubal, Praveen Krishnan, Punit~Singh Koura, Puxin Xu, Qing He, Qingxiao Dong, Ragavan Srinivasan, Raj Ganapathy, Ramon Calderer, Ricardo~Silveira Cabral, Robert Stojnic, Roberta Raileanu, Rohit Girdhar, Rohit Patel, Romain Sauvestre, Ronnie Polidoro, Roshan Sumbaly, Ross Taylor, Ruan Silva, Rui Hou, Rui Wang, Saghar Hosseini, Sahana Chennabasappa, Sanjay Singh, Sean Bell, Seohyun~Sonia Kim, Sergey Edunov, Shaoliang Nie, Sharan Narang, Sharath Raparthy, Sheng Shen, Shengye Wan, Shruti Bhosale, Shun Zhang, Simon Vandenhende, Soumya Batra, Spencer Whitman, Sten Sootla, Stephane Collot, Suchin Gururangan, Sydney Borodinsky, Tamar Herman, Tara Fowler, Tarek Sheasha, Thomas Georgiou, Thomas Scialom, Tobias Speckbacher, Todor Mihaylov, Tong Xiao, Ujjwal Karn, Vedanuj Goswami, Vibhor Gupta, Vignesh Ramanathan, Viktor Kerkez, Vincent Gonguet, Virginie Do, Vish Vogeti, Vladan Petrovic, Weiwei Chu,
  Wenhan Xiong, Wenyin Fu, Whitney Meers, Xavier Martinet, Xiaodong Wang, Xiaoqing~Ellen Tan, Xinfeng Xie, Xuchao Jia, Xuewei Wang, Yaelle Goldschlag, Yashesh Gaur, Yasmine Babaei, Yi~Wen, Yiwen Song, Yuchen Zhang, Yue Li, Yuning Mao, Zacharie~Delpierre Coudert, Zheng Yan, Zhengxing Chen, Zoe Papakipos, Aaditya Singh, Aaron Grattafiori, Abha Jain, Adam Kelsey, Adam Shajnfeld, Adithya Gangidi, Adolfo Victoria, Ahuva Goldstand, Ajay Menon, Ajay Sharma, Alex Boesenberg, Alex Vaughan, Alexei Baevski, Allie Feinstein, Amanda Kallet, Amit Sangani, Anam Yunus, Andrei Lupu, Andres Alvarado, Andrew Caples, Andrew Gu, Andrew Ho, Andrew Poulton, Andrew Ryan, Ankit Ramchandani, Annie Franco, Aparajita Saraf, Arkabandhu Chowdhury, Ashley Gabriel, Ashwin Bharambe, Assaf Eisenman, Azadeh Yazdan, Beau James, Ben Maurer, Benjamin Leonhardi, Bernie Huang, Beth Loyd, Beto~De Paola, Bhargavi Paranjape, Bing Liu, Bo~Wu, Boyu Ni, Braden Hancock, Bram Wasti, Brandon Spence, Brani Stojkovic, Brian Gamido, Britt Montalvo, Carl
  Parker, Carly Burton, Catalina Mejia, Changhan Wang, Changkyu Kim, Chao Zhou, Chester Hu, Ching-Hsiang Chu, Chris Cai, Chris Tindal, Christoph Feichtenhofer, Damon Civin, Dana Beaty, Daniel Kreymer, Daniel Li, Danny Wyatt, David Adkins, David Xu, Davide Testuggine, Delia David, Devi Parikh, Diana Liskovich, Didem Foss, Dingkang Wang, Duc Le, Dustin Holland, Edward Dowling, Eissa Jamil, Elaine Montgomery, Eleonora Presani, Emily Hahn, Emily Wood, Erik Brinkman, Esteban Arcaute, Evan Dunbar, Evan Smothers, Fei Sun, Felix Kreuk, Feng Tian, Firat Ozgenel, Francesco Caggioni, Francisco Guzmán, Frank Kanayet, Frank Seide, Gabriela~Medina Florez, Gabriella Schwarz, Gada Badeer, Georgia Swee, Gil Halpern, Govind Thattai, Grant Herman, Grigory Sizov, Guangyi, Zhang, Guna Lakshminarayanan, Hamid Shojanazeri, Han Zou, Hannah Wang, Hanwen Zha, Haroun Habeeb, Harrison Rudolph, Helen Suk, Henry Aspegren, Hunter Goldman, Ibrahim Damlaj, Igor Molybog, Igor Tufanov, Irina-Elena Veliche, Itai Gat, Jake Weissman, James
  Geboski, James Kohli, Japhet Asher, Jean-Baptiste Gaya, Jeff Marcus, Jeff Tang, Jennifer Chan, Jenny Zhen, Jeremy Reizenstein, Jeremy Teboul, Jessica Zhong, Jian Jin, Jingyi Yang, Joe Cummings, Jon Carvill, Jon Shepard, Jonathan McPhie, Jonathan Torres, Josh Ginsburg, Junjie Wang, Kai Wu, Kam~Hou U, Karan Saxena, Karthik Prasad, Kartikay Khandelwal, Katayoun Zand, Kathy Matosich, Kaushik Veeraraghavan, Kelly Michelena, Keqian Li, Kun Huang, Kunal Chawla, Kushal Lakhotia, Kyle Huang, Lailin Chen, Lakshya Garg, Lavender A, Leandro Silva, Lee Bell, Lei Zhang, Liangpeng Guo, Licheng Yu, Liron Moshkovich, Luca Wehrstedt, Madian Khabsa, Manav Avalani, Manish Bhatt, Maria Tsimpoukelli, Martynas Mankus, Matan Hasson, Matthew Lennie, Matthias Reso, Maxim Groshev, Maxim Naumov, Maya Lathi, Meghan Keneally, Michael~L. Seltzer, Michal Valko, Michelle Restrepo, Mihir Patel, Mik Vyatskov, Mikayel Samvelyan, Mike Clark, Mike Macey, Mike Wang, Miquel~Jubert Hermoso, Mo~Metanat, Mohammad Rastegari, Munish Bansal, Nandhini
  Santhanam, Natascha Parks, Natasha White, Navyata Bawa, Nayan Singhal, Nick Egebo, Nicolas Usunier, Nikolay~Pavlovich Laptev, Ning Dong, Ning Zhang, Norman Cheng, Oleg Chernoguz, Olivia Hart, Omkar Salpekar, Ozlem Kalinli, Parkin Kent, Parth Parekh, Paul Saab, Pavan Balaji, Pedro Rittner, Philip Bontrager, Pierre Roux, Piotr Dollar, Polina Zvyagina, Prashant Ratanchandani, Pritish Yuvraj, Qian Liang, Rachad Alao, Rachel Rodriguez, Rafi Ayub, Raghotham Murthy, Raghu Nayani, Rahul Mitra, Raymond Li, Rebekkah Hogan, Robin Battey, Rocky Wang, Rohan Maheswari, Russ Howes, Ruty Rinott, Sai~Jayesh Bondu, Samyak Datta, Sara Chugh, Sara Hunt, Sargun Dhillon, Sasha Sidorov, Satadru Pan, Saurabh Verma, Seiji Yamamoto, Sharadh Ramaswamy, Shaun Lindsay, Shaun Lindsay, Sheng Feng, Shenghao Lin, Shengxin~Cindy Zha, Shiva Shankar, Shuqiang Zhang, Shuqiang Zhang, Sinong Wang, Sneha Agarwal, Soji Sajuyigbe, Soumith Chintala, Stephanie Max, Stephen Chen, Steve Kehoe, Steve Satterfield, Sudarshan Govindaprasad, Sumit Gupta,
  Sungmin Cho, Sunny Virk, Suraj Subramanian, Sy~Choudhury, Sydney Goldman, Tal Remez, Tamar Glaser, Tamara Best, Thilo Kohler, Thomas Robinson, Tianhe Li, Tianjun Zhang, Tim Matthews, Timothy Chou, Tzook Shaked, Varun Vontimitta, Victoria Ajayi, Victoria Montanez, Vijai Mohan, Vinay~Satish Kumar, Vishal Mangla, Vítor Albiero, Vlad Ionescu, Vlad Poenaru, Vlad~Tiberiu Mihailescu, Vladimir Ivanov, Wei Li, Wenchen Wang, Wenwen Jiang, Wes Bouaziz, Will Constable, Xiaocheng Tang, Xiaofang Wang, Xiaojian Wu, Xiaolan Wang, Xide Xia, Xilun Wu, Xinbo Gao, Yanjun Chen, Ye~Hu, Ye~Jia, Ye~Qi, Yenda Li, Yilin Zhang, Ying Zhang, Yossi Adi, Youngjin Nam, Yu, Wang, Yuchen Hao, Yundi Qian, Yuzi He, Zach Rait, Zachary DeVito, Zef Rosnbrick, Zhaoduo Wen, Zhenyu Yang, and Zhiwei Zhao.
\newblock The llama 3 herd of models, 2024.
\newblock URL \url{https://arxiv.org/abs/2407.21783}.

\bibitem[Elsken et~al.(2019)Elsken, Metzen, and Hutter]{Elsken2019NeuralAS}
Thomas Elsken, Jan~Hendrik Metzen, and Frank Hutter.
\newblock Neural architecture search: A survey.
\newblock \emph{ArXiv}, abs/1808.05377, 2019.

\bibitem[Esser et~al.(2020)Esser, Rombach, and Ommer]{esser2020taming}
Patrick Esser, Robin Rombach, and Björn Ommer.
\newblock Taming transformers for high-resolution image synthesis, 2020.

\bibitem[Gao et~al.(2024)Gao, Wang, Chen, Liu, Wu, Chen, and Li]{gao2024parameterefficientfinetuningdiscretefourier}
Ziqi Gao, Qichao Wang, Aochuan Chen, Zijing Liu, Bingzhe Wu, Liang Chen, and Jia Li.
\newblock Parameter-efficient fine-tuning with discrete fourier transform, 2024.

\bibitem[Gou et~al.(2021)Gou, Yu, Maybank, and Tao]{Gou-2021}
Jianping Gou, Baosheng Yu, Stephen~J. Maybank, and Dacheng Tao.
\newblock Knowledge distillation: A survey.
\newblock \emph{International Journal of Computer Vision}, 129\penalty0 (6):\penalty0 1789--1819, mar 2021.

\bibitem[He et~al.(2016)He, Zhang, Ren, and Sun]{He2016DeepRL}
Kaiming He, X.~Zhang, Shaoqing Ren, and Jian Sun.
\newblock Deep residual learning for image recognition.
\newblock \emph{2016 IEEE Conference on Computer Vision and Pattern Recognition (CVPR)}, pp.\  770--778, 2016.

\bibitem[Huang et~al.(2023)Huang, Liu, Lin, Pang, Du, and Lin]{huang2023lorahub}
Chengsong Huang, Qian Liu, Bill~Yuchen Lin, Tianyu Pang, Chao Du, and Min Lin.
\newblock Lorahub: Efficient cross-task generalization via dynamic lora composition, 2023.

\bibitem[Knyazev et~al.(2021)Knyazev, Drozdzal, Taylor, and Romero-Soriano]{Knyazev2021ParameterPF}
Boris Knyazev, Michal Drozdzal, Graham~W. Taylor, and Adriana Romero-Soriano.
\newblock Parameter prediction for unseen deep architectures.
\newblock \emph{ArXiv}, abs/2110.13100, 2021.

\bibitem[Peebles et~al.(2022)Peebles, Radosavovic, Brooks, Efros, and Malik]{Peebles2022LearningTL}
William~S. Peebles, Ilija Radosavovic, Tim Brooks, Alexei~A. Efros, and Jitendra Malik.
\newblock Learning to learn with generative models of neural network checkpoints.
\newblock \emph{ArXiv}, abs/2209.12892, 2022.

\bibitem[Radford et~al.(2019)Radford, Wu, Child, Luan, Amodei, Sutskever, et~al.]{radford2019language}
Alec Radford, Jeffrey Wu, Rewon Child, David Luan, Dario Amodei, Ilya Sutskever, et~al.
\newblock Language models are unsupervised multitask learners.
\newblock \emph{OpenAI blog}, 1\penalty0 (8):\penalty0 9, 2019.

\bibitem[Radford et~al.(2021)Radford, Kim, Hallacy, Ramesh, Goh, Agarwal, Sastry, Askell, Mishkin, Clark, Krueger, and Sutskever]{radford2021learningtransferablevisualmodels}
Alec Radford, Jong~Wook Kim, Chris Hallacy, Aditya Ramesh, Gabriel Goh, Sandhini Agarwal, Girish Sastry, Amanda Askell, Pamela Mishkin, Jack Clark, Gretchen Krueger, and Ilya Sutskever.
\newblock Learning transferable visual models from natural language supervision, 2021.

\bibitem[Ratzlaff \& Fuxin(2019)Ratzlaff and Fuxin]{pmlr-v97-ratzlaff19a}
Neale Ratzlaff and Li~Fuxin.
\newblock {H}yper{GAN}: A generative model for diverse, performant neural networks.
\newblock In Kamalika Chaudhuri and Ruslan Salakhutdinov (eds.), \emph{Proceedings of the 36th International Conference on Machine Learning}, volume~97 of \emph{Proceedings of Machine Learning Research}, pp.\  5361--5369. PMLR, 09--15 Jun 2019.

\bibitem[Russakovsky et~al.(2015)Russakovsky, Deng, Su, Krause, Satheesh, Ma, Huang, Karpathy, Khosla, Bernstein, Berg, and Fei-Fei]{ILSVRC15}
Olga Russakovsky, Jia Deng, Hao Su, Jonathan Krause, Sanjeev Satheesh, Sean Ma, Zhiheng Huang, Andrej Karpathy, Aditya Khosla, Michael Bernstein, Alexander~C. Berg, and Li~Fei-Fei.
\newblock {ImageNet Large Scale Visual Recognition Challenge}.
\newblock \emph{International Journal of Computer Vision (IJCV)}, 115\penalty0 (3):\penalty0 211--252, 2015.
\newblock \doi{10.1007/s11263-015-0816-y}.

\bibitem[Sch{\"u}rholt et~al.(2022{\natexlab{a}})Sch{\"u}rholt, Knyazev, i~Nieto, and Borth]{Schrholt2022HyperRepresentationsAG}
Konstantin Sch{\"u}rholt, Boris Knyazev, Xavier~Gir{\'o} i~Nieto, and Damian Borth.
\newblock Hyper-representations as generative models: Sampling unseen neural network weights.
\newblock In \emph{Advances in Neural Information Processing Systems}, 2022{\natexlab{a}}.

\bibitem[Sch{\"u}rholt et~al.(2022{\natexlab{b}})Sch{\"u}rholt, Knyazev, i~Nieto, and Borth]{schrholt2022hyperrepresentation}
Konstantin Sch{\"u}rholt, Boris Knyazev, Xavier~Gir{\'o} i~Nieto, and Damian Borth.
\newblock Hyper-representation for pre-training and transfer learning.
\newblock In \emph{First Workshop on Pre-training: Perspectives, Pitfalls, and Paths Forward at ICML 2022}, 2022{\natexlab{b}}.

\bibitem[Sch{\"u}rholt et~al.(2022{\natexlab{c}})Sch{\"u}rholt, Taskiran, Knyazev, i~Nieto, and Borth]{schurholt2022model}
Konstantin Sch{\"u}rholt, Diyar Taskiran, Boris Knyazev, Xavier~Gir{\'o} i~Nieto, and Damian Borth.
\newblock Model zoos: A dataset of diverse populations of neural network models.
\newblock In \emph{Thirty-sixth Conference on Neural Information Processing Systems Datasets and Benchmarks Track}, 2022{\natexlab{c}}.

\bibitem[Sch{"u}rholt et~al.(2024)Sch{"u}rholt, Mahoney, and Borth]{schuerholt2024sane}
Konstantin Sch{"u}rholt, Michael~W. Mahoney, and Damian Borth.
\newblock Towards scalable and versatile weight space learning.
\newblock In \emph{Proceedings of the 41st International Conference on Machine Learning (ICML)}, 2024.

\bibitem[Schürholt et~al.(2021)Schürholt, Kostadinov, and Borth]{schurholt-self-supervised-2021}
Konstantin Schürholt, Dimche Kostadinov, and Damian Borth.
\newblock Self-supervised representation learning on neural network weights for model characteristic prediction.
\newblock In \emph{Advances in Neural Information Processing Systems (NeurIPS 2021)}, Sydney, Australia, 2021.

\bibitem[Shu et~al.(2021)Shu, Kou, Cao, Wang, and Long]{pmlr-v139-shu21b}
Yang Shu, Zhi Kou, Zhangjie Cao, Jianmin Wang, and Mingsheng Long.
\newblock Zoo-tuning: Adaptive transfer from a zoo of models.
\newblock In Marina Meila and Tong Zhang (eds.), \emph{Proceedings of the 38th International Conference on Machine Learning}, volume 139 of \emph{Proceedings of Machine Learning Research}, pp.\  9626--9637. PMLR, 18--24 Jul 2021.

\bibitem[Soro et~al.(2024)Soro, Andreis, Lee, Chong, Hutter, and Hwang]{soro2024diffusionbased}
Bedionita Soro, Bruno Andreis, Hayeon Lee, Song Chong, Frank Hutter, and Sung~Ju Hwang.
\newblock Diffusion-based neural network weights generation, 2024.

\bibitem[Stanley \& Miikkulainen(2002)Stanley and Miikkulainen]{Stanley2002EvolvingNN}
Kenneth~O. Stanley and Risto Miikkulainen.
\newblock Evolving neural networks through augmenting topologies.
\newblock \emph{Evolutionary Computation}, 10:\penalty0 99--127, 2002.

\bibitem[Tang et~al.(2024)Tang, Lv, Zhang, Wu, and Kuang]{tang2024modelgptunleashingllmscapabilities}
Zihao Tang, Zheqi Lv, Shengyu Zhang, Fei Wu, and Kun Kuang.
\newblock Modelgpt: Unleashing llm's capabilities for tailored model generation, 2024.

\bibitem[Ullah et~al.(2022)Ullah, Carrion, Escalera, Guyon, Huisman, Mohr, van Rijn, Sun, Vanschoren, and Vu]{meta-album-2022}
Ihsan Ullah, Dustin Carrion, Sergio Escalera, Isabelle~M Guyon, Mike Huisman, Felix Mohr, Jan~N van Rijn, Haozhe Sun, Joaquin Vanschoren, and Phan~Anh Vu.
\newblock Meta-album: Multi-domain meta-dataset for few-shot image classification.
\newblock In \emph{Thirty-sixth Conference on Neural Information Processing Systems Datasets and Benchmarks Track}, 2022.
\newblock URL \url{https://meta-album.github.io/}.

\bibitem[van~den Oord et~al.(2017)van~den Oord, Vinyals, and Kavukcuoglu]{Oord2017NeuralDR}
A{\"a}ron van~den Oord, Oriol Vinyals, and Koray Kavukcuoglu.
\newblock Neural discrete representation learning.
\newblock In \emph{NIPS}, 2017.

\bibitem[Wang et~al.(2024)Wang, Xu, Zhou, Zang, Darrell, Liu, and You]{wang2024neural}
Kai Wang, Zhaopan Xu, Yukun Zhou, Zelin Zang, Trevor Darrell, Zhuang Liu, and Yang You.
\newblock Neural network diffusion, 2024.

\bibitem[Wightman(2019)]{rw2019timm}
Ross Wightman.
\newblock Pytorch image models.
\newblock \url{https://github.com/rwightman/pytorch-image-models}, 2019.

\bibitem[Yang et~al.(2021)Yang, Xiao, Shen, Jiang, Hu, Zhang, and Peng]{Yang2021ASO}
Jian Yang, Gang Xiao, Yulong Shen, Wei Jiang, Xinyu Hu, Ying Zhang, and Jinghui Peng.
\newblock A survey of knowledge enhanced pre-trained models.
\newblock \emph{ArXiv}, abs/2110.00269, 2021.

\bibitem[Zhang et~al.(2019)Zhang, Ren, and Urtasun]{Zhang2019GraphHF}
Chris Zhang, Mengye Ren, and Raquel Urtasun.
\newblock Graph hypernetworks for neural architecture search.
\newblock \emph{ArXiv}, abs/1810.05749, 2019.

\bibitem[Zhao et~al.(2024)Zhao, Gan, Wang, Hu, Shen, Yang, Kuang, and Wu]{zhao2024retrievalaugmentedmixtureloraexperts}
Ziyu Zhao, Leilei Gan, Guoyin Wang, Yuwei Hu, Tao Shen, Hongxia Yang, Kun Kuang, and Fei Wu.
\newblock Retrieval-augmented mixture of lora experts for uploadable machine learning, 2024.

\bibitem[Zhmoginov et~al.(2022)Zhmoginov, Sandler, and Vladymyrov]{Zhmoginov2022HyperTransformerMG}
Andrey Zhmoginov, Mark Sandler, and Max Vladymyrov.
\newblock Hypertransformer: Model generation for supervised and semi-supervised few-shot learning.
\newblock \emph{ArXiv}, abs/2201.04182, 2022.

\end{thebibliography}
\bibliographystyle{iclr2025_conference}

\appendix

\clearpage
\setcounter{page}{1}

\textbf{Limitations}: A key limitation of our method is its reliance on training with a large, diverse set of pretrained models—a comprehensive public repository of such models does not yet exist. However, this challenge is increasingly mitigated by the growing availability of models from repositories like Hugging Face and by efficient fine-tuning techniques such as LoRA.
\section{Overview}

\subsubsection{Vectorizing Neural Network Parameters}\label{vector}

To enable our generative mapping function $H$ to learn from diverse pretrained models, we introduce a standardized parameter vectorization scheme that transforms weights and biases into a uniform vector representation. For a network with $L$ layers, fully connected layers are vectorized by reshaping the weight matrix $\theta^{(l)} \in \mathbb{R}^{d_{l-1} \times d_l}$ into $\operatorname{vec}(\theta^{(l)}) \in \mathbb{R}^{d_{l-1}d_l}$ and appending the bias $b^{(l)} \in \mathbb{R}^{d_l}$, yielding $d_{l-1}d_l + d_l$ elements. Similarly, convolutional layers with kernel $\theta^{(l)} \in \mathbb{R}^{k_h \times k_w \times c_{\text{in}} \times c_{\text{out}}}$ are flattened to $\operatorname{vec}(\theta^{(l)}) \in \mathbb{R}^{k_hk_wc_{\text{in}}c_{\text{out}}}$ and concatenated with the bias $b^{(l)} \in \mathbb{R}^{c_{\text{out}}}$ to produce $k_hk_wc_{\text{in}}c_{\text{out}} + c_{\text{out}}$ elements. We consider two aggregation strategies: an architecture-wise vectorization that concatenates all layers into a single vector $\Theta = \bigoplus_{l=1}^L\left[\operatorname{vec}(\theta^{(l)}) \oplus b^{(l)}\right]$, and a layer-wise encoding that preserves each layer’s representation. 

\subsection{Model Overview}{\label{modelover}}
Our VQVAE model is a modified implementation based on the VQGAN codebase. Table~\ref{tab:vqvae_tokenmixer} summarizes the architecture details, while Table~\ref{tab:model_configuration} provides an overview of the generative model along with its parameter counts. We optimize the model using Adam with a learning rate of $1\times10^{-4}$ and employ a cyclical temperature schedule for the Gumbel-Softmax, annealing the temperature from $1$ to $1\times10^{-4}$. During both training and inference, when image dataset encoding is unnecessary, the model conditions solely on the architecture; otherwise, it leverages both architectural and dataset embeddings. We also illustrate an example template~\ref{tab:model_configuration} for experimental results in Figure~\ref{bar_cross_res}, where only architecture embeddings are used since all architectures were pretrained on the same dataset.

\begin{table}[h!]
\centering
\resizebox{1.0\textwidth}{!}{
\begin{tabular}{|c|c|c|c|}
\hline
\textbf{Index} & \textbf{Name} & \textbf{Type} & \textbf{Parameters} \\
\hline
0 & \texttt{encoder} & \texttt{Encoder} & 197\,\text{M} \\
1 & \texttt{decoder} & \texttt{Decoder} & 198\,\text{M} \\
2 & \texttt{loss} & \texttt{VQLoss} & 0 \\
3 & \texttt{quantize} & \texttt{GumbelQuantize} & 132\,\text{K} \\
4 & \texttt{quant\_conv} & \texttt{Conv2d} & 4.2\,\text{K} \\
5 & \texttt{post\_quant\_conv} & \texttt{Conv2d} & 4.2\,\text{K} \\
\hline
\end{tabular}
}
\caption{Pretrained weights configuration for the VQVAE model. The table details the layer-wise parameters of the VQVAE model, highlighting its encoder, decoder, and quantization components optimized for downstream tasks.}
\label{tab:vqvae_tokenmixer}
\end{table}

\begin{table}[h!]
\centering
\resizebox{1.0\textwidth}{!}{
\begin{tabular}{|c|c|c|c|}
\hline
\textbf{Index} & \textbf{Name} & \textbf{Type} & \textbf{Parameters} \\
\hline
0 & \texttt{first\_stage\_model} & \texttt{GumbelVQNoDisc} & 48.5\,\text{M} \\
1 & \texttt{cond\_stage\_model} & \texttt{CondStage} & 524\,\text{K} \\
2 & \texttt{ArchCond} & \texttt{Identity} & 0 \\
3 & \texttt{transformer} & \texttt{GPT} & 27.8\,\text{M} \\
\hline
\end{tabular}
}
\caption{Layer-wise configuration of the model, including the first-stage model, conditional stage model, and transformer components. The table specifies the type and parameter count of each component. This table shows the structure of the model used for the experiments in Section~\ref{zoo}}
\label{tab:model_configuration}
\end{table}

\begin{remark}
Instruction: You are a codebook generator. Input: Generate the full codebook of length 512 for ResNet with: num\_blocks = [6, 7, 6]. Output:[995, 66, 750, 391, 809, 830,...].
\label{tmp}
\end{remark}

\section{Additional Results and Tables}\label{add}
\begin{algorithm}[t]
\caption{Autoregressive Parameter Generation}
\label{alg:param_gen}
\begin{algorithmic}[1]
\Require Architecture encoding $e_\mathcal{A}$, dataset encoding $e_\mathcal{D}$, parameter length $L$, chunk size $K$
\Ensure Architecture parameters $\theta_\mathcal{A} \in \mathbb{R}^L$
\State $k \gets \lceil L/K \rceil$ \Comment{Number of chunks}
\If{$kl \leq N_\text{max}$}
    \State $\mathbf{s} \gets \mathcal{G}(e_\mathcal{A}, e_\mathcal{D})$ \Comment{Single-pass generation}
    \For{$i \gets 1$ to $k$}
        \State $\mathbf{s}_i \gets \mathbf{s}_{(i-1)l:il}$ \Comment{Split into chunks}
        \State $\theta_i \gets \mathbf{D}(\mathbf{s}_i)$ \Comment{Decode chunks}
    \EndFor
\Else
    \State $\mathbf{s}^{(1)} \gets \mathcal{G}(e_\mathcal{A}, e_\mathcal{D})$ \Comment{Initial generation}
    \For{$j \gets 2$ to $\lceil kl/N_\text{max} \rceil$}
        \State $\mathbf{s}^{(j)} \gets \mathcal{G}(e_\mathcal{A}, e_\mathcal{D}, \mathbf{s}^{(j-1)}_\text{ctx})$
    \EndFor
    \State $\mathbf{s}_\text{full} \gets \text{concat}(\mathbf{s}^{(1)}, \ldots, \mathbf{s}^{(j)})_{1:kl}$
    \For{$i \gets 1$ to $k$}
        \State $\mathbf{s}_i \gets (\mathbf{s}_\text{full})_{(i-1)l:il}$
        \State $\theta_i \gets \mathbf{D}(\mathbf{s}_i)$
    \EndFor
\EndIf
\State $\theta_\mathcal{A} \gets \mathcal{F}([\theta_1; \ldots; \theta_k])$ \Comment{Flatten chunks}
\State \Return $\theta_\mathcal{A}$
\end{algorithmic}
\end{algorithm}

\begin{table}[ht]

\centering
\resizebox{0.5\textwidth}{!}{
\begin{tabular}{llc}
\toprule
\textbf{Domain} & \textbf{Original Dataset Name} & \textbf{Number of Classes} \\
\midrule
\multirow{3}{*}{Large Animals} 
  & Birds                        & 315 \\
  & Dogs                         & 120 \\
  & Animals with Attributes      & 50  \\
\midrule
\multirow{3}{*}{Small Animals} 
  & Plankton                     & 102 \\
  & Insects 2                    & 102 \\
  & Insects                      & 117 \\
\midrule
\multirow{3}{*}{Plants} 
  & Flowers                      & 102 \\
  & PlantNet                     & 25  \\
  & Fungi                        & 25  \\
\midrule
\multirow{3}{*}{Plant Diseases} 
  & Plant Village                & 38  \\
  & Medicinal Leaf               & 26  \\
  & PlantDoc                     & 27  \\
\midrule
\multirow{3}{*}{Microscopy} 
  & Bacteria                     & 33  \\
  & PanNuke                      & 19  \\
  & Subcellular Human Protein    & 21  \\
\midrule
\multirow{3}{*}{Remote Sensing} 
  & RESISC                       & 45  \\
  & RSICB                        & 45  \\
  & RSD                          & 43  \\
\midrule
\multirow{3}{*}{Vehicles} 
  & Cars                         & 196 \\
  & Airplanes                    & 21  \\
  & Boats                        & 26  \\
\midrule
\multirow{3}{*}{Manufacturing} 
  & Textures                     & 64  \\
  & Textures DTD                 & 47  \\
  & Textures ALOT                & 250 \\
\midrule
\multirow{3}{*}{Human Actions} 
  & Sports                       & 73  \\
  & Stanford 40 Actions          & 40  \\
  & MPII Human Pose              & 29  \\
\midrule
\multirow{3}{*}{OCR} 
  & Omniprint-MD-mix             & 706 \\
  & Omniprint-MD-5-bis           & 706 \\
  & Omniprint-MD-6               & 703 \\
\bottomrule
\end{tabular}
}
\caption{Diverse pretrained datasets grouped by domain, showcasing their variety in classes for cross-dataset adaptation tasks.}
\label{tab:datasets}
\end{table}

\begin{figure}[h!]
\centering
\includegraphics[width=1.0\linewidth]{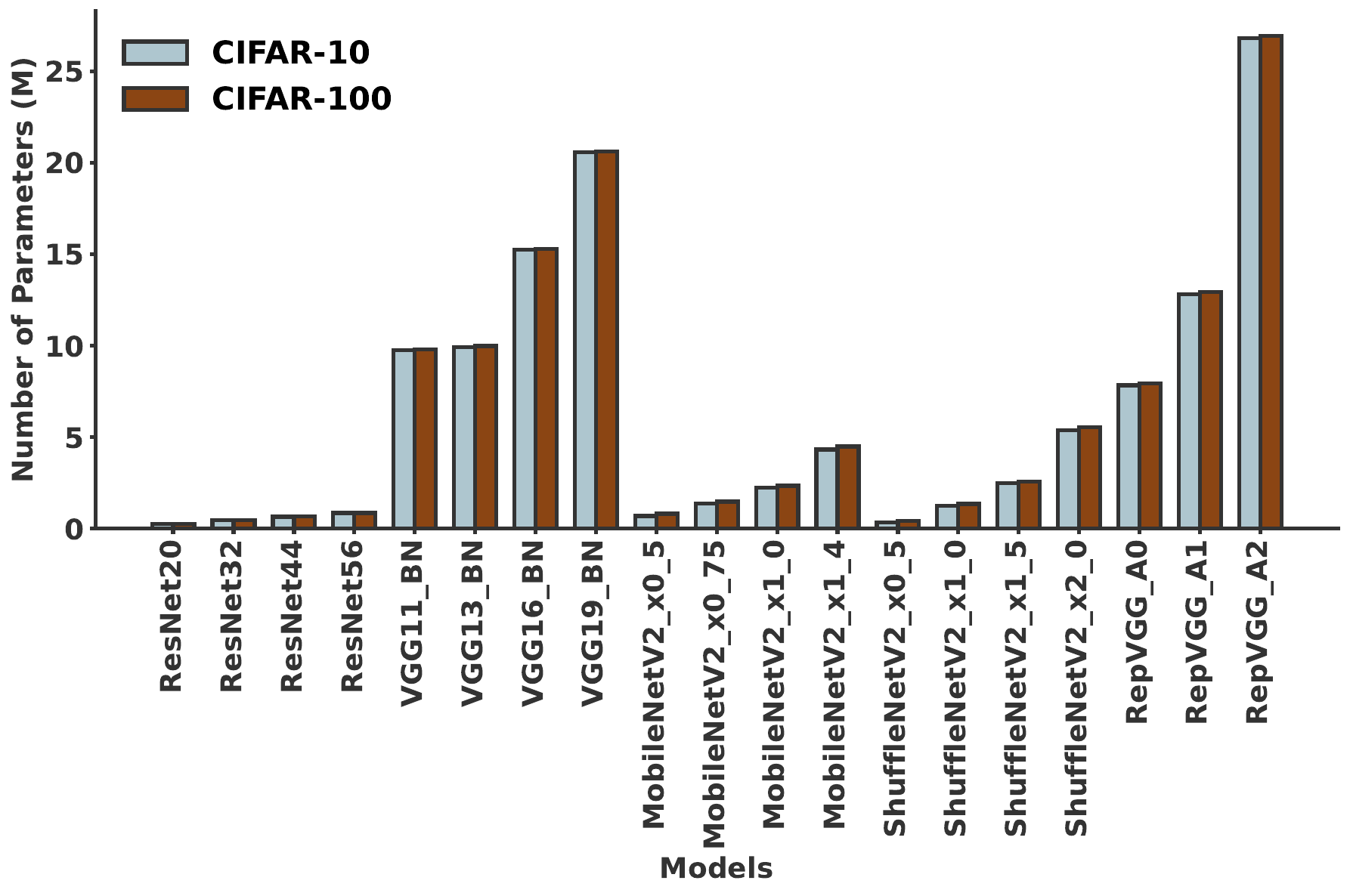}
    \caption[]{\small Parameters distribution of diverse architectures pretrained on CIFAR-10 and CIFAR100 all jointly encoded by \ourmethod}
    \vspace{-0.15in}
    \label{bar_param}
\end{figure}

In this section, we provide a comprehensive report of the experimental results such as those presented in section~\ref{multiples}. The primary objective of these results is to demonstrate the ability to encode diverse architectures across a variety of datasets while maintaining retrieval performance comparable to the pretrained models. Table \ref{tab:model_comparison} showcases representative results for different architectures, along with their corresponding parameter counts. The conditioning has been applied to both architectural configurations and datasets. Additionally, we include example configuration samples used for instruction encoding in Table \ref{tab:architecture_configurations}.

\begin{table}[h!]
\centering
\caption{Examples of architecture configurations for various neural network families.}
\resizebox{1.0\textwidth}{!}{
\begin{tabular}{ll}
\toprule
Architecture Name &                                      Configuration \\
\midrule
vgg11\_bn & architecture: A, layers: [64, 'M', 128, 'M', 256, 256, 'M', 512, 512, 'M', 512, 512, 'M'], batch\_norm: true \\
mobilenetv2\_x0\_5 & width\_multiplier: 0.5, final\_layer\_channels: 640 \\
repvgg\_a0 & width\_multiplier: [0.75, 0.75, 0.75, 2.5] \\
shufflenetv2\_x0\_5 & stages\_repeats: [4, 8, 4], stages\_out\_channels: [24, 48, 96, 192, 1024] \\
resnet20 & layers: [3, 3, 3] \\
\bottomrule
\end{tabular}
}
\label{tab:architecture_configurations}
\end{table}

\section{Related Work}
\textbf{Neural Network Parameter Generation}
Parameter generation for neural networks has evolved along two main trajectories: hypernetwork-based generation and generative hyper-representation learning. Hypernetwork approaches generate weights from scratch~\citep{Stanley2002EvolvingNN, pmlr-v97-ratzlaff19a, Deutsch2018GeneratingNN} , with recent advances in graph-hypernetworks~\citep{Zhang2019GraphHF, Knyazev2021ParameterPF} and transformer architectures~\citep{Zhmoginov2022HyperTransformerMG}. However, these methods primarily serve as initialization techniques requiring subsequent optimization.
In contrast, generative hyper-representation learning focuses on modeling distributions of pretrained weights. Recent work has explored adaptive weight transfer~\citep{pmlr-v139-shu21b}, learning distribution of pretrained weights ~\citep{schrholt2022hyperrepresentation}, and diffusion-based generation~\citep{Peebles2022LearningTL} which demonstrated promising results. Our current work extend these work to autoregressive generation preserving inter-layer relation in the network.

\textbf{Applications and Implications}
Weight generation, particularly from pretrained distributions, offers several key advantages in modern deep learning such as enabling efficient model compression and serving for large language models by enabling the generation of task-specific adaptations ~\citep{huang2023lorahub, loraland2024} or act as parameters retrieval-based systems similar to ~\citet{zhao2024retrievalaugmentedmixtureloraexperts}. Second, it enhances transfer learning by enabling task-adaptive sampling ~\citep{tang2024modelgptunleashingllmscapabilities}, streamlining model adaptation. 

\section{Ablation Study}\label{ablat}

\paragraph{Full Model vs Layer-Wise Sampling} 
We compare two parameter generation approaches: full model-wise encoding and layer-wise encoding. In layer-wise encoding, we assemble parameters from each layer into separate training datasets, applying chunking and padding per layer. While both methods perform well on in-distribution parameters, layer-wise encoding shows superior generalization to novel architectures, suggesting better adaptability and robustness.

\paragraph{Impact of Chunking Strategy}
We evaluate the effect of chunking network weights versus using complete weight vectors. Using pretrained weights from PyTorchHub\footnote{\url{https://github.com/chenyaofo/pytorch-cifar-models}}, we assess ResNet56 and MobileNetV2 on CIFAR-10 and CIFAR-100. Results in Table~\ref{tab:ch_fr} show that while chunking offers no significant advantage for medium-sized architectures, it becomes crucial for larger models where chunk-free approaches struggle to maintain performance.

\paragraph{LLM-based Parameter Generation}
We investigate whether instruction-tuned LLMs can generate neural network parameters directly. Using LLaMA-3.2-1B-Instruct with LoRA fine-tuning and GPT-2 trained on sequence-to-sequence codebook generation, we find mixed results. While LLaMA-3.2-1B accurately generates initial tokens, it struggles with longer sequences. Similarly, GPT-2 with top-k sampling (k=1) successfully matches pretrained codebooks for small parameter sets but degrades significantly beyond 1024 dimensions. These results indicate that LLMs can generate VQVAE codebook parameters for small models, but scalability remains a significant challenge.

\begin{table}[t]
\caption{Performance comparison of conditional sampling with chunk-based and chunk-free approach. \textbf{C-10} and \textbf{C-100} refer to respectively CIFAR-10 and CIFAR-100 datasets}
\label{tab:ch_fr}
\vspace{-0.2in}
\begin{center}
\begin{small}
\begin{sc}
\resizebox{1.0\columnwidth}{!}{
\begin{tabular}{lcccccccr}
\toprule
 &  & \multicolumn{2}{c}{pretrained} & \multicolumn{2}{c}{chunk-free} & \multicolumn{2}{c}{chunk-based} &  \\
 models& \#Params & top-1 & top-5 & top-1 & top-5  & top-1 & top5 & Datasets \\

\midrule
resnet-56 & 860026 & 94.37 & \textbf{99.83} & \textbf{94.40}$\pm$0.01 & 99.82$\pm$0.00  & 94.19$\pm$0.01 & 99.87$\pm$ 0.00 & C-10 \\
Resnet-56 & 865876  & \textbf{72.63} & 91.94 & 72.45$\pm$0.01 & \textbf{91.96$\pm$0.01} & 72.41 $\pm$ 0.2 & 91.80 $\pm$0.03 & C-100 \\
mobilenetv2\_x0\_5 & 834324 & 70.88 & 91.72 & \textbf{71.14}$\pm$0.01 & 92.48$\pm$0.04 & 70.93 $\pm$0.00 & \textbf{92.55 $\pm$0.02 } & C-100\\
\bottomrule
\end{tabular}
}
\end{sc}
\end{small}
\end{center}
\vspace{-0.2in}
\end{table}

\subsection{Neural Network Codebook Generation Using Chat Models}\label{chat:sec}

We conducted several attempts to generate neural network codebooks using large language models (LLMs) optimized for chat-based interactions. Our investigation with LLaMA-3.2-1B revealed significant challenges in adapting chat models to generate neural network parameters without pretraining on the specific codebook data. For instance, training GPT-2-small in a sequence-to-sequence fashion with the codebook, followed by instruction tuning, enabled the model to successfully generate a correct codebook in 96.6

However, simply applying LoRA tuning to an instruction-tuned LLaMA model resulted in the generation of short and inconsistent codebooks, often with mixed or incomplete outputs. These findings highlight the limitations of current chat models in directly producing neural network parameters without substantial fine-tuning or pretraining on task-specific data.

In future work, we aim to explore this problem more comprehensively, focusing on improving parameter generation capabilities with chat models. Most of our experiments to date utilized GPT-2 variants, and we plan to expand this investigation with other LLM architectures.

\begin{table*}[t]
\caption{Performance of weights generation from diverses models pretrained on CIFAR-10 and CIFAR-100 datasets}
\label{tab:model_comparison}
\centering
\begin{tabular}{l|ccc|ccc}
\toprule
\multirow{2}{*}{Model} & \multicolumn{3}{c|}{CIFAR-10} & \multicolumn{3}{c}{CIFAR-100} \\
& IGPG & Pretrained & Params (M) & IGPG & Pretrained & Params (M) \\
\midrule
MobileNetV2 (×0.5) & 93.16 & 93.12 & 0.70 & 71.05 & 71.14 & 0.82 \\
MobileNetV2 (×0.75) & 94.16 & 94.08 & 1.37 & 74.15 & 74.06 & 1.48 \\
MobileNetV2 (×1.0) & 93.98 & 94.05 & 2.24 & 74.41 & 74.30 & 2.35 \\
MobileNetV2 (×1.4) & 94.26 & 94.22 & 4.33 & 75.97 & 76.29 & 4.50 \\
\textbf{Average (MobileNetV2)} & \textbf{93.89} & \textbf{93.87} & \textbf{2.16} & \textbf{73.90} & \textbf{73.95} & \textbf{2.29} \\
\midrule
RepVGG-A0 & 94.44 & 94.47 & 7.84 & 75.32 & 75.29 & 7.96 \\
RepVGG-A1 & 94.93 & 94.93 & 12.82 & 76.51 & 76.45 & 12.94 \\
RepVGG-A2 & 95.36 & 95.27 & 26.82 & 77.52 & 77.50 & 26.94 \\
\textbf{Average (RepVGG)} & \textbf{94.91} & \textbf{94.89} & \textbf{15.83} & \textbf{76.45} & \textbf{76.41} & \textbf{15.95} \\
\midrule
ResNet-20 & 92.47 & 92.59 & 0.27 & 68.92 & 68.84 & 0.28 \\
ResNet-32 & 93.47 & 93.53 & 0.47 & 69.98 & 70.15 & 0.47 \\
ResNet-44 & 93.92 & 94.01 & 0.66 & 71.39 & 71.64 & 0.67 \\
ResNet-56 & 94.56 & 94.37 & 0.86 & 72.43 & 72.60 & 0.86 \\
\textbf{Average (ResNet)} & \textbf{93.61} & \textbf{93.63} & \textbf{0.57} & \textbf{70.68} & \textbf{70.81} & \textbf{0.57} \\
\midrule
ShuffleNetV2 (×0.5) & 90.61 & 90.65 & 0.35 & 67.90 & 67.80 & 0.44 \\
ShuffleNetV2 (×1.0) & 92.75 & 93.30 & 1.26 & 72.55 & 72.59 & 1.36 \\
ShuffleNetV2 (×1.5) & 93.58 & 93.57 & 2.49 & 74.22 & 74.22 & 2.58 \\
ShuffleNetV2 (×2.0) & 94.02 & 93.99 & 5.37 & 75.39 & 75.49 & 5.55 \\
\textbf{Average (ShuffleNetV2)} & \textbf{92.74} & \textbf{92.88} & \textbf{2.37} & \textbf{72.52} & \textbf{72.53} & \textbf{2.48} \\
\midrule
VGG11-BN & 92.85 & 92.78 & 9.76 & 70.76 & 70.78 & 9.80 \\
VGG13-BN & 94.01 & 94.00 & 9.94 & 74.40 & 74.62 & 9.99 \\
VGG16-BN & 94.19 & 94.15 & 15.25 & 73.98 & 74.00 & 15.30 \\
VGG19-BN & 94.09 & 93.91 & 20.57 & 73.82 & 73.87 & 20.61 \\
\textbf{Average (VGG)} & \textbf{93.79} & \textbf{93.71} & \textbf{13.88} & \textbf{73.24} & \textbf{73.32} & \textbf{13.93} \\
\midrule
\textbf{Global Average (All Models)} & \textbf{93.74} & \textbf{93.76} & \textbf{6.96} & \textbf{73.36} & \textbf{73.42} & \textbf{7.00} \\
\bottomrule
\end{tabular}
\end{table*}

\end{document}